\documentclass[10pt,journal,compsoc]{IEEEtran}

\ifCLASSOPTIONcompsoc
  \usepackage[nocompress]{cite}
\else
  \usepackage{cite}
\fi

\usepackage{amsmath}
\usepackage{times} 
\usepackage{amssymb,longtable,calc}
\usepackage{mathptmx}       
\usepackage{graphicx}
\usepackage{subfigure}
\usepackage[textwidth=2cm,colorinlistoftodos]{todonotes}
\usepackage{xspace}
\usepackage{latexsym}
\usepackage{amsmath,amssymb,graphicx,xspace}
\usepackage{times}   
\usepackage{multirow}
\usepackage{color,comment,rotating,subfigure,url,soul} 
\usepackage{tikz}
\usetikzlibrary{arrows,shadows,shapes,backgrounds,decorations,snakes,fit}
\usepackage{algorithm}

\usepackage[algo2e]{algorithm2e}

\usepackage{amsfonts}

\makeatletter

\renewcommand\paragraph{\@startsection{paragraph}{4}{\z@}%
                                    {3.25ex \@plus1ex \@minus.2ex}%
                                    {-1em}%
                                    {\normalfont\normalsize\bfseries}}
\makeatother

\usepackage[top=1in, bottom=1in, left=1.25in, right=1.25in]{geometry}

\usepackage{todonotes}
\usepackage{enumerate}
\usepackage{bm}

\DeclareMathOperator{\E}{\mathbb{E}}

\DeclareMathOperator{\Dkl}{D_{KL}}
 
\newcommand{\talph}{$\alpha$}

\newcommand{\bmx}{\bm{x}} 
\newcommand{\bmz}{\bm{z}} 
\newcommand{\x}{x} 
\newcommand{\z}{z} 

\newcommand{\lz}{\xi}			
\newcommand{\bmlz}{\bm{\lz}}	
\newcommand{\bmlphi}{\bm{\phi}} 

\newcommand{\glmd}{\gamma}
\newcommand{\gtht}{\theta}
\newcommand{\bmlambda}{\bm{\lambda}}

\newcommand{\ff}{\tilde{f}}

\newcommand{\subparagraph}[1]{ \textit{#1}}

\ifCLASSINFOpdf
\else
\fi

\begin{document}
\title{Advances in Variational Inference}

\author{Cheng  Zhang  ~ \IEEEmembership{Member,~IEEE,} ~~
        Judith B\"utepage~ \IEEEmembership{Member,~IEEE,}\protect \\
       Hedvig Kjellstr\"om ~\IEEEmembership{Member,~IEEE,}~~
        Stephan Mandt~ \IEEEmembership{Member,~IEEE}
\IEEEcompsocitemizethanks{\IEEEcompsocthanksitem Cheng Zhang is with Microsoft Research, Cambridge, UK, and prior to that was with Disney Research,  USA and the KTH Royal Institute of Technology, Sweden.
 \protect\\
E-mail: cheng.zhang@microsoft.com
\IEEEcompsocthanksitem  Judith B\"utepage is with the KTH Royal Institute of Technology, Sweden. Part of her contribution in this work was done during her internship at Disney Research, USA.  \protect\\
E-mail: butepage@kth.se
\IEEEcompsocthanksitem  Hedvig Kjellstr\"om is with the KTH Royal Institute of Technology, Sweden. \protect\\
E-mail: hedvig@kth.se
\IEEEcompsocthanksitem   Stephan Mandt is with the University of California, Irvine, and prior to that was with Disney Research, USA. \protect\\
E-mail: stephan.mandt@gmail.com
}
}

\markboth{}%
{Shell \MakeLowercase{\textit{et al.}}: Bare Advanced Demo of IEEEtran.cls for IEEE Computer Society Journals}

\IEEEtitleabstractindextext{%
\begin{abstract}
Many modern unsupervised or semi-supervised machine learning algorithms rely on Bayesian probabilistic models. These models are usually intractable and thus require approximate inference. Variational inference (VI) lets us approximate a high-dimensional Bayesian posterior with a simpler variational distribution by solving an optimization problem. This approach has been successfully applied to various models and large-scale applications.
In this review, we give an overview of recent trends in variational inference. 
We first introduce standard mean field variational inference, then review recent advances   
focusing on the following aspects: (a) \emph{scalable} VI, which includes stochastic approximations, (b) \emph{generic} VI, which extends the applicability of VI to a large class of otherwise intractable models, such as non-conjugate models, (c) \emph{accurate} VI, which includes variational models beyond the mean field approximation or  with atypical divergences, and (d) \emph{amortized} VI, which implements the inference over local latent variables with inference networks. Finally, we provide a summary of promising future research directions. 
\end{abstract}

\begin{IEEEkeywords}
Variational Inference, Approximate Bayesian Inference, Reparameterization Gradients, Structured Variational Approximations, Scalable Inference, Inference Networks.
\end{IEEEkeywords}}

\maketitle

\IEEEdisplaynontitleabstractindextext

%
\IEEEpeerreviewmaketitle
\section{Introduction}

Bayesian inference has become a crucial component of machine learning.  It allows us to systematically reason about parameter uncertainty. The central object of interest in Bayesian inference is the posterior distribution of model parameters given observations. 
This review focuses on \emph{variational inference} (VI): a methodology 
that makes Bayesian inference computationally efficient and scalable to large data sets.


Bayesian machine learning frequently relies on probabilistic latent variable models, such as Gaussian mixture models, Hidden Markov models, Latent Dirichlet Allocation, stochastic block models, and Bayesian deep learning architectures. Computing the exact Bayesian posterior requires to sum or integrate over all latent variables, which can be in the millions or billions for complex models and large-scale applications. Exact inference is therefore typically intractable in these models, and approximations are needed. 

The central idea of VI is to approximate
the model posterior by a simpler distribution. To this end, one minimizes the Kullback-Leibler divergence between the posterior and the approximating distribution. This approach circumvents computing intractable normalization constants. It only requires knowledge of the joint distribution of the observations and the latent variables. This methodology along with its recent refinements will be reviewed in this paper.

Within the field of approximate Bayesian inference, VI falls into the class of optimization-based approaches  \cite{bishop06,hennig2011approximate}. This class also contains methods such as loopy belief propagation \cite{murphy1999loopy} and
expectation propagation (EP) \cite{minka2001expectation}. On the contrary, Markov Chain Monte Carlo (MCMC)  approaches rely on sampling \cite{heinrich08,porteous2008fast,brooks2011handbook}.  By construction, MCMC is often unbiased, and thus converges to the true posterior in the limit, but it can be slow to converge. 
Optimization-based methods, on the other hand, are often faster  but may suffer from oversimplified posterior approximations \cite{bishop06,wainwright2008graphical}. 
In recent years, there has been considerable progress in both fields \cite{blei2017variational, angelino2016patterns}, and in particular on bridging the gap between these methods \cite{ahn2012bayesian,korattikara2015bayesian,salimans14,ranganath2014black,mandt2016variational}. In fact, recent progress in scalable VI partly relies on fusing optimization-based and sampling-based methods. While this review focuses on VI, readers interested in EP and MCMC are referred to, e.g., \cite{angelino2016patterns} and \cite{seeger2005expectation}.   

The origins of VI date back to the 1980s. Mean field methods, for instance, have their origins in statistical physics, where they played a prominent role in the statistical mechanics of spin glasses \cite{mezard1990spin, parisi1988statistical}.  Early applications of variational methods also include the study of neural networks  \cite{opper1996mean, peterson1987mean}. The latter work inspired the computer science community of the 1990s to adopt variational methods in the context of probabilistic graphical models \cite{saul1996mean, jaakkola1996fast,jordan1999introduction,opper2001advanced} .

In recent years, several factors have driven a renewed interest in variational methods. The modern versions of VI differ significantly from earlier formulations. Firstly, the availability of large datasets triggered the interest in \emph{scalable} approaches, e.g., based on stochastic gradient descent \cite{bottou2010large,hoffman13}. Secondly, classical VI is limited to  conditionally conjugate exponential family models, a restricted class of models described in \cite{wainwright2008graphical, hoffman13}. In contrast, black box VI algorithms \cite{jaakkola1998improving,jordan1999introduction,ranganath2014black} and probabilistic programs facilitate \emph{generic} VI, making it applicable to a range of complicated models. Thirdly, this generalization has spurred research on more \emph{accurate} variational approximations, such as alternative divergence measures \cite{minka05divergence,li2016renyi,zhang2017perturbative} and structured variational families \cite{ranganath2016hierarchical}. Finally, \emph{amortized} inference employs complex functions  such as neural networks to predict variational distributions conditioned on data points, rendering VI an important component of modern Bayesian deep learning architectures such as variational autoencoders. In this work, we discuss important papers concerned with each of these four aspects.

While several excellent reviews of VI exist, we believe that our focus on recent developments in \emph{scalable, generic, accurate} and  \emph{amortized} VI goes beyond those efforts. Both \cite{jordan1999introduction} and \cite{opper2001advanced} date back to the early $2000$s and do not cover the developments of recent years. Similarly, \cite{wainwright2008graphical} is an excellent resource, especially regarding structured approximations and the information geometrical aspects of VI. However, it was  published prior to the widespread use of stochastic methods in VI. Among recent introductions, \cite{blei2017variational} contains many examples, empirical comparisons, and explicit model calculations but focuses less on recent developments while \cite{angelino2016patterns} mainly focuses on scalable MCMC. 
Our review concentrates on the advances of the last 10 years prior to the publication of this paper. Complementing previous reviews, we skip example calculations to focus on a more  exhaustive survey of the recent literature. For readers who are new to the field, we recommend to read Chapter 10 on approximate inference in \cite{bishop06} as a preparation.

We survey the trends and developments in VI in a self-contained manner.  Section \ref{sec:ingredients} covers basic concepts, such as variational distributions and the evidence lower bound. 
In the succeeding sections, we concentrate on recent advances and identify four main research directions: scalable VI (Section  \ref{sec:scalability}), generic VI (Section  \ref{sec:generalizability}), accurate VI (Section  \ref{sec:accuracy}), and amortized VI (Section  \ref{sec:DAIforDL}). 
We finalize the review with a discussion (Section  \ref{sec:link}) and concluding remarks (Section  \ref{sec:conclusions}).

\section{Variational Inference}
\label{sec:ingredients}

We begin this review with a brief tutorial on variational inference, presenting the mathematical foundations of this procedure and explaining the basic mean-field approximation. 

The generative process is specified by observations $\bmx$, as well as latent variables $\bmz$ and a joint distribution $p(\bmx,\bmz)$. We use bold font to explicitly indicate sets of variables, i.e. $\bmz = \{ z_1, z_2, \cdots, z_N \}$, where $N$ is the total number of latent variables and $\bmx = \{ x_1, x_2, \cdots, x_M \}$, where $M$ is the total number of observations in the dataset.
The variational distribution  $q(\bmz; \bmlambda)$  is defined over the latent variables $\bmz$ and has variational parameters $\bmlambda = \{\lambda_1, \lambda_2, \cdots, \lambda_N \}$. \footnote{Note that the number variational parameter is not necessary the same as the number of latent variables. Latent variables can be shared among multiple data points, and individual data points can have multiple latent variables. }

\subsection{Inference as Optimization}
\label{sec:KLDivergence}

The central object of interest in Bayesian statistics is the posterior distribution of latent variables given observations:
\begin{align}
p(\bmz|\bmx) = \frac{p(\bmx,\bmz)}{\int p(\bmx,\bmz) d \bmz}.
\end{align}
For most models, this integral is high dimensional, thus computing the normalization term is intractable. 

Instead of computing the posterior normalization, the basic idea of VI is to approximate the posterior with a simpler distribution. This involves a \emph{variational} distribution $q(\bmz;\bmlambda)$, characterized by a set of \emph{variational} parameters $\bmlambda$. These parameters are tuned to obtain the best matching.  
Finally, 
the optimized variational distribution is taken as a proxy for the posterior. In this way, VI turns Bayesian inference into an optimization problem over variational parameters.

For two distributions $p(z)$ and $q(z)$, a \emph{divergence} $D(q(z)|| p(z))$ measures the difference between the distributions, such that $D(q(z)|| p(z)) \geq 0$ and $D(q(z)|| p(z)) = 0$ only for $q(z)=p(z)$. VI amounts to minimizing a divergence between the variational distribution and the posterior. We show below that this does not require knowing the posterior normalization.

While various divergence measures exist \cite{amari1985differential,minka05divergence, amari2009divergence,stein1972bound}, the most commonly used divergence is the Kullback-Leibler (KL) divergence \cite{kullback1951information,bishop06}, which is also referred to as relative entropy or information gain:
\begin{equation}
\Dkl(q(z)|| p(z) ) = - \int q(z) \log \frac{p(z)}{q(z)} dz. \label{eq:KLdivergence}
\end{equation}
As seen in Eq. \ref{eq:KLdivergence}, the KL divergence is asymmetric; $\Dkl(q(z)||p(z)) \neq \Dkl(p(z)||q(z))$. Depending on the ordering, we obtain two different approximate inference methods. As we show below, VI employs $
\Dkl( q(\bmz;\bmlambda) || p(\bmz|\bmx) )   =  -\E _{q(\bmz;\bmlambda)}   \left[ \log \frac{p(\bmz|\bmx)}{q(\bmz;\bmlambda)} \right]$.  
In contrast and as an aside, expectation propagation (EP) \cite{minka2001expectation} optimizes $\Dkl( p(\bmz|\bmx) || q(\bmz)) $ for local moment matching, 
which is not reviewed in this paper\footnote{We refer the readers to the EP roadmap for more information about advancements of EP. \url{https://tminka.github.io/papers/ep/roadmap.html}}.

\subsection{Variational Objective}
\label{sec:VIbasic}

Classical VI aims at determining a variational distribution $q(\bmz; \bm{\lambda} )$ that is as close as possible to the posterior $p(\bmz|\bmx)$, measured in terms of the KL divergence. 
Minimizing the KL divergence to zero would guarantee that the variational distribution matches the exact posterior. However, in practice
this is rarely possible: the variational distribution is usually under-parameterized and thus not sufficiently flexible to capture the full complexity of the true posterior.

As follows, we will show that minimizing the KL divergence is equivalent to maximizing a related quantity, the
\textit{Evidence Lower BOund (ELBO)} $\mathcal{L}$. The ELBO is a lower bound on the log marginal probability of the data and 
can be derived from $\log p(\bmx)$ using Jensen's inequality as follows:
\begin{equation}
\begin{aligned}
\log p(\bmx) &= \log \int p(\bmx,\bmz) d\bmz = \log \int \frac{p(\bmx,\bmz) q(\bmz; \bm{\lambda})}{q(\bmz; \bm{\lambda})} d\bmz \\
&= \log \E _{q(\bmz; \bm{\lambda})}  \left[  \frac{p(\bmx,\bmz)}{q(\bmz; \bm{\lambda})}  \right]\\ 
&\ge   \E_{q(\bmz; \bm{\lambda})} \left[ \log \frac{p(\bmx,\bmz)}{q(\bmz; \bm{\lambda} )}  \right] \equiv \mathcal{L(\bm{\lambda})}.
\label{eq:ELBO}
\end{aligned}
\end{equation}
It can be shown (see Appendix \ref{sec:ELBO_KL}) that the difference between the true log marginal probability of the data and the ELBO is the KL divergence between the variational distribution and the posterior: 
\begin{equation}
\log p(\bmx) =  \mathcal{L(\bm{\lambda})} + \Dkl( q || p )
\label{eq:ELBO_KL}
\end{equation}
Thus, maximizing the ELBO is equivalent to minimizing the KL divergence between $q$ and $p$, where $q$ and $p$ replace $q(\bmz; \bm{\lambda})$ and $p(\bmz| \bmx)$ for the sake of brevity. Since the ELBO is a conservative estimate of this marginal, 
it is sometimes taken as an estimate of how well the model fits the data. The ELBO can also be used for model selection.

In traditional VI, computing the ELBO amounts to analytically solving the expectations over $q$. This restricts the class of tractable models
to the so-called conditionally conjugate exponential family (see Appendix \ref{sec:ExpFam} and   \cite{wainwright2008graphical}). 
For an example calculation to derive the ELBO analytically for a mixture of Gaussians, we refer to \cite{blei2017variational}. Section~\ref{sec:generalizability} introduces modern alternatives to computing these expectations.

\subsection{Mean Field Variational Inference}
\label{sec:MFVI}

There is a tradeoff in choosing $q(\bmz;\bmlambda)$ expressive enough to approximate the posterior well, and simple enough to lead to a tractable approximation \cite{bishop06}.  
A common choice is a fully factorized distribution, also called mean field distribution. 
A mean field approximation assumes that all latent variables are independent, which simplifies derivations. However, this independence assumption also leads to less accurate results especially when the true posterior variables are highly dependent. Section~\ref{sec:accuracy}
 discusses a more expressive class of variational distributions.

Mean Field Variational Inference (MFVI) has its origins in the mean field theory of physics \cite{opper2001advanced}.
In this approximation, the variational distribution factorizes, and each factor is governed by its own variational parameter:
\begin{equation}
q(\bmz; \bm{\lambda}) = \prod_{i=1}^N q(z_i; \lambda_i).
\label{eq:MF}
\end{equation}
For notational simplicity, we omit the variational parameters $\bmlambda$ for the remainder of this section. 
We now review how to maximize the ELBO $\mathcal{L}$, defined in Eq. \ref{eq:ELBO}, under a mean field assumption. 

A fully factorized variational distribution allows one to optimize $\mathcal{L}$ via simple iterative updates. To see this, we focus on updating the variational parameter $\lambda_j$ associated with latent variable $z_j$.
Inserting the mean field distribution into Eq. \ref{eq:ELBO} allows us to express the ELBO as follows:
\begin{equation}
\begin{aligned}
\label{eq:One-d-ELBO}
\mathcal{L} = &\int q(z_j) \E_{q(\bmz_{\neg j}) } \left[\log p( z_j , \bmx | \bmz_{\neg j})\right] dz_j \\
&- \int q(z_j) \log  q(z_j)  d z_j +  c_j.
\end{aligned}
\end{equation}
Above, $\bmz_{\neg j}$ indicates the set $\bmz$  excluding $z_j$. The constant $c_j$ contains all terms that are constant with respect to  $z_j$, such as the entropy term associated with $\bmz_{\neg j}$. We have thus separated the full expectation into an inner expectation over $\bmz_{\neg j}$, and an outer expectation over $z_j$. 
 
Eq. \ref{eq:One-d-ELBO} assumes the form of a negative KL divergence, 
which is maximized for variable $j$ by
\begin{equation}
\begin{aligned}
\log \ q^*(z_j) =  \E_{q(\bmz_{\neg j})} \left[ \log p(z_j | \bmz_{\neg j}, \bmx )\right] + const.
\end{aligned}
\end{equation}
Exponentiating and normalizing this result yields:
\begin{equation}
\begin{aligned}
q^*(z_j)& \propto \exp( \E_{q(\bmz_{\neg j})}  \left[ \log p(z_j | \bmz_{\neg j}, \bmx )\right] )\\
& \propto \exp( \E_{q(\bmz_{\neg j})}  \left[ \log p(\bmz, \bmx )\right] )
\label{eq:update}
\end{aligned}
\end{equation}
Using Eq. \ref{eq:update}, the variational distribution can be updated iteratively for each latent variable until convergence. Similar updates also form the basis for the variational message passing algorithm \cite{winn2005variational} (Appendix \ref{sec:VMP}).
 
For more details on the mean field approximation and its geometrical interpretation we refer the reader to \cite{bishop06} and \cite{wainwright2008graphical}.

\subsection{Beyond Vanilla Variational Inference}
Classical MFVI has historically played an important role, however, it is limited in multiple ways when it comes to modern applications. One of the challenges is to scale VI to big datasets. This will be addressed in Section \ref{sec:scalability}, where we show that VI can be combined with stochastic optimization and distributed computing to achieve this goal. 
Big datasets and fast algorithms allow for more sophisticated models. In order to make VI tractable for this modern class of models (in particular for so-called non-conjugate ones), Section \ref{sec:generalizability} describes methods that make VI both easier to use and more generic. 
Furthermore, certain models and applications require more accurate inference techniques, such as improved variational approximations and tighter bounds. 
A popular stream of research is concerned with alternative divergence measures beyond the KL divergence, and will be reviewed in Section~\ref{sec:accuracy}, where we also review non-mean field variational approximations. 
Finally, we describe in Section~\ref{sec:DAIforDL} how neural networks can be used to amortize the estimiation of certain local latent variables. This leads to a significant speedup for many models and bridges the gap between Bayesian inference and modern representation learning.

\section{Scalable Variational Inference}
\label{sec:scalability}

In this section, we survey scalable VI. Big datasets raise new challenges for the computational feasibility of Bayesian algorithms, making scalable inference techniques essential. We begin by reviewing stochastic variational inference (SVI) in Section \ref{sec:SVI}, which uses stochastic gradient descent (SGD) to scale VI to large datasets. Section \ref{sec:FactConvergence} discusses practical aspects of SVI, such as adaptive learning rates and variance reduction. Further approaches to improve on the scalability of VI are discussed in Section \ref{sec:modelspecific}; these include sparse inference, collapsed inference, and distributed inference.

This section follows the general model structure of global and local hidden variables, assumed in \cite{hoffman13}. Fig. \ref{fig:generativemodel} depicts the corresponding graphical model where the latent variable $\bmz = \{\gtht, \bmlz\}$ includes local (per data point) variables $\bmlz =\{ \xi_1, ... , \xi_M \}$  and global variable $\gtht$. Similarly, the variational parameters are given by $\bmlambda = \{\glmd, \bmlphi \}$, where the variational parameter $\glmd$ corresponds to  the global latent variable, and $\bmlphi$ denotes the set of local variational parameters. Furthermore, the model depends on hyperparameters $\alpha$. The  mini-batch size is denoted by $S$.

\subsection{Stochastic Variational Inference}
\label{sec:SVI}

We showed that VI frames Bayesian inference as an optimization problem. For many models of interest, the variational objective has a special structure, namely, it is the sum over contributions from all $M$ individual data points. Problems of this type can be solved efficiently using stochastic optimization \cite{robbins1951stochastic,bottou2010large}. Stochastic Variational Inference amounts to applying stochastic optimization to the objective function encountered in VI \cite{honkela2003line,hoffman13,hoffman2010online,wang2011online}, thereby scaling VI to  very large datasets. Using stochastic optimization in the context of VI was proposed in \cite{sato2001online,honkela2003line,hoffman13}. 
We follow the conventions of \cite{hoffman13} which presents SVI for models of the conditionally conjugate exponential family class. 

\begin{figure}
\centering   
\scalebox{1.0}{\pgfdeclarelayer{background}
\pgfdeclarelayer{foreground}
\pgfsetlayers{background,main,foreground}

\begin{tikzpicture}

\tikzstyle{surround} = [thick,draw=black,rounded corners=1mm]
\tikzstyle{scalarnode} = [circle, draw, fill=white!11,  
    text width=1.2em, text badly centered, inner sep=2.5pt]
    \tikzstyle{Vnode} = [circle, draw, fill=black, inner sep=2.5pt]
\tikzstyle{arrowline} = [draw,color=black, -latex]
    
    \node [Vnode] at (-1.5,0) (A)   {};
    \node[]at (-1.8,0) (Anote)   {$\alpha$};
    \node [scalarnode] at (0,0) (O)   {$\gtht$};
    \node [scalarnode] at (0, -1.5) (Z) {$\lz$};
    \node [scalarnode, fill=black!30 ] at (1.5, -1.5) (X) {$x$};
    \node[surround, inner sep = .3cm] (f_N) [fit = (Z)(X) ] {};
    \node [] at (1.9, -1.9) (M) {M};
    \path [arrowline]  (A) to (O);
    \path [arrowline]  (O) to (Z);
    \path [arrowline] (Z) to (X);
    \path[arrowline]  (O) to (X);
\end{tikzpicture}}
\caption{A graphical model of the observations $\bmx$ that depend on underlying local hidden factors $\bmlz$ and global parameters $\gtht$. We use $\bmz =\{\gtht, \bmlz\}$ to represent  all latent variables. $M$ is the number of the data points. $N$ is the number of the latent variables. \vspace{-5pt}} 
\label{fig:generativemodel}
\end{figure}
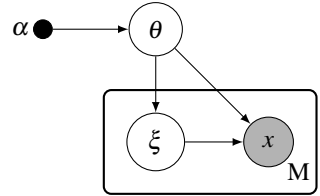

The ELBO of the general graphical model shown in Fig. \ref{fig:generativemodel} has the following form:
\begin{align}
\mathcal{L} &= \E_q [ \log p(\gtht|\alpha) - \log q(\gtht|\gamma)] +  \label{Eq:ELBOwithTheta} \\ 
& \sum_{i=1}^M \E_q \big[ \log p(\lz_i | \gtht) + \log  p(x_i| \lz_i, \gtht) - \log q(\lz_i|\phi_i)\big].  \nonumber
\end{align}
We assume that the variational distribution is given by $q(\bmlz,\gtht) = q(\gtht|\gamma) \prod_{i=1}^M q(\lz_i|\phi_i)$. Here, we also assume that the expectations in Eq.~\ref{Eq:ELBOwithTheta} are analytically tractable, yielding a closed-form objective.

Eq.~\ref{Eq:ELBOwithTheta} could be optimized by coordinate descent (Section~\ref{sec:ingredients}), or gradient descent on the ELBO. 
In both cases, every iteration or gradient step scales with $M$, and is therefore expensive for large data. In contrast,
SVI solves this problem in the spirit of stochastic gradient descent ~\cite{bottou2010large}.
In every iteration, one randomly selects mini-batches of size $S$ to obtain a stochastic estimate of the ELBO $\hat{\mathcal{L}}$,
\begin{align}
\label{Eq:ELBOSVI}
\hat{\mathcal{L}} & =  \E_q [ \log p(\gtht|\alpha) - \log q(\gtht|\gamma)]   +  \\ \nonumber
\frac{M}{S}\sum_{s=1}^S & \E_q \big[ \log p(\lz_{i_s} | \gtht) + \log  p(\x_{i_s}| \lz_{i_s}, \gtht) - \log q(\lz_{i_s}|\phi_{i_s})\big], 
\end{align}
where $i_s$ is the variable index from the mini-batch. Then, the gradient of Eq.~\ref{Eq:ELBOSVI} is computed, which gives a noisy estimator of the direction of steepest ascent of the true ELBO. 

\tikzstyle{unobserved} = [draw, circle, node distance=1.5cm,text centered, text width=1.5em]
\tikzstyle{observed} = [draw, circle,fill=gray!40, node distance=1.5cm,text centered, text width=1.5em]

An important result of \cite{hoffman13} is that using \emph{natural} gradients instead of standard gradients in SVI simplifies the variational updates for models in the conditionally conjugate exponential family. 
Natural gradients, first studied in \cite{amari1998natural} and introduced to VI in 
\cite{sato2001online,honkela2007natural,honkela2010approximate}, 
take the information geometry of the model into account. They are obtained by pre-multiplying the gradient with the inverse Fisher information matrix. While we skip further discussions for brevity, interested readers are referred to Appendix \ref{sec:svi_naturalgradient} and \cite{hoffman13}.  Recent advances in this direction include \cite{hensman2012fast,lin2018variational, khan2018fast}.

SVI requires the same conditions for convergence as regular SGD. The minibatch indices $i_s$ must be drawn uniformly at random. The size $S$ of the minibatch satisfies  $1 \leq S \ll M$. Larger values of $S$ reduce the variance of the stochastic natural gradient. When $S=M$, SVI reduces to traditional batch VI when the learning rate is set to $1$. However, computational savings are only obtained for $S \ll M$.
 The optimal choice of $S$ emerges from a trade-off between the computational overhead associated with processing a mini-batch, such as performing inference over global parameters (favoring larger mini-batches which have lower gradient noise, allowing larger learning rates), and the cost of iterating over local parameters in the mini-batch (favoring small mini-batches).  Additionally, this tradeoff is also affected by memory structures in modern hardware such as GPUs. The learning rate $\rho_t$ needs to decrease with iterations $t$, satisfying the Robbins-Monro conditions
$\sum _ {t=1} ^ {\infty} \rho _t = \infty$ and $\sum _ {t=1} ^ {\infty} \rho _t ^2 < \infty$. This guarantees that every point in the parameter space can be reached, while the gradient noise decreases quickly enough to ensure convergence \cite{robbins1951stochastic}.

Sometimes SVI is referred to as online VI  \cite{hoffman2010online,wang2011online}. These methods are equivalent under the assumptions that the volume of the data $M$ is known. In streaming applications, the mini-batches arrive sequentially from a data source, but the SVI updates are the same.
However, when $M$ is unknown, it is unclear how to set the scale parameter $M/S$ in Eq. \ref{Eq:ELBOSVI}. To this end,  \cite{mcinerney2015population} introduces the concept of the population posterior which depends on the unknown size of the dataset. This concept allows one to apply online VI with respect to the expected ELBO over the population. 

Stochastic gradient methods
have been adapted to various settings, such as gamma processes \cite{knowles2015stochastic} and variational autoencoders \cite{kingma2014stochastic}. In recent years, most advancements in VI have been developed relying on a SVI scheme. In the following, we  detail how to further adapt SVI to accelerate convergence.

\subsection{Tricks of the Trade for SVI}
\label{sec:FactConvergence}

The convergence speed of SGD, forming the basis of SVI, depends on the variance of the gradient estimates. Smaller gradient noise allows for larger learning rates and leads to faster convergence. This section covers tricks of the trade in the context of SVI, such as adaptive learning rates and variance reduction. Some of these approaches are generally applicable in SGD.

\paragraph{Adaptive Learning Rate and Mini-batch Size} 
The speed of convergence is influenced by  the choice of the learning rate and the mini-batch size \cite{balles2016coupling,friedlander2012hybrid}. Due to the law of large numbers, increasing the mini-batch size reduces the stochastic gradient noise \cite{friedlander2012hybrid}, allowing larger learning rates. To accelerate the learning procedure, one can either optimally adapt the mini-batch size for a given learning rate, or optimally adjust the learning rate to a fixed mini-batch size.

We begin by discussing learning rate adaptation.
In each iteration, the empirical gradient variance can guide the adaptation of the learning rate which is inversely proportional to the gradient noise. Popular optimization methods that make use of this idea include  RMSProp \cite{tieleman2012lecture}, AdaGrad \cite{duchi2011adaptive}, AdaDelta \cite{zeiler2012adadelta} and  Adam \cite{kingma2014adam}. These methods are not specific to SVI but are frequently used in this context; for more details we refer interested readers to \cite{Goodfellow-et-al-2016}. 

\cite{ranganath2013adaptive} first introduced adaptive learning rates for the global variational parameter $\gamma$ in SVI,
where the optimal learning rate was shown to satisfy
\begin{equation}
\rho _{t} ^* = \frac{ ( \glmd_t^* - \glmd_t) ^T ( \glmd_t^* - \glmd_t)  }{ ( \glmd_t^* - \glmd_t) ^T ( \glmd_t^* - \glmd_t) + tr(\Sigma)}.
\label{eq:adaptrate}
\end{equation}
Above, $\glmd_t^*$ denotes the optimal global variational parameter, and $\glmd_t$ the current estimate. $\Sigma$ is the covariance matrix of the variational parameter in this mini-batch. Since $\glmd_t^*$ is unknown, \cite{ranganath2013adaptive} showed how to estimate the optimal learning rate in an online fashion. 

Instead of adapting the learning rate, the mini-batch size can be adapted while keeping the learning rate fixed. This achieves similar effects  \cite{balles2016coupling,byrd2012sample,de2017big, tan2017stochastic}. In order to decrease the SGD variance, \cite{balles2016coupling} proposed to choose the mini-batch size proportionally to the value of the objective function relative to its optimum. In practice, the estimated gradient noise covariance and the magnitude of the gradient are used to estimate the optimal mini-batch size.

\paragraph{Variance Reduction}
In addition to controlling the optimization path through the learning rate and mini-batch size, we can reduce the variance, thereby enabling larger gradient steps. Variance reduction is often employed in SVI to achieve faster convergence. 
As follows, we summarize the literature on how to accomplish this goal via \emph{control variates}, \emph{non-uniform sampling}, and \emph{other approaches}. 

\vspace{10pt}
\subparagraph{Control Variates.}
A control variate is a stochastic term that can be added to the stochastic gradient such that its expectation remains the same, but its variance is reduced \cite{boyle1977options}. 
A control variate needs to be correlated with the stochastic gradient, and easy to compute.
Using control variates for variance reduction is common in Monte Carlo simulation and stochastic optimization \cite{ross06,wang2013variance}. Several authors have suggested the use of control variates in the context of SVI \cite{paisley2012variational, wang2013variance, johnson2013accelerating, ranganath2014black}. 

As a prominent example, we discuss the stochastic variance reduced gradient (SVRG) method \cite{johnson2013accelerating}. In 
SVRG, one constructs a control
variate which takes advantage of previous gradients from all data points, and one exploits that gradients along the optimization path are correlated.
The standard stochastic gradient update  $\glmd_{t+1} = \glmd_{t} - \rho_t (\nabla \hat{\mathcal{L}} (\glmd_t) )$ is replaced by
\begin{equation}
\glmd_{t+1} = \glmd_{t} - \rho_t (\nabla \hat{\mathcal{L}} (\glmd_t) - \nabla \hat{\mathcal{L}} (\tilde{\glmd}) + \tilde{\mu} ).
\end{equation}
 $\hat{\mathcal{L}}$ indicates the estimated objective (here the negative ELBO) based on the current set of mini-batch indices, $\tilde{\glmd}$ is a snapshot of $\glmd$ after every $m$ iterations, and $\tilde{\mu}$ is the batch gradient computed over all the data points, $\tilde{\mu} = \nabla \mathcal{L} (\tilde{\glmd})$. Since $- \nabla \hat{\mathcal{L}} (\tilde{\glmd}) + \tilde{\mu}$ has expectation zero, it is a control variate.
 
SVRG requires a full pass through the dataset every $m_{th}$ iteration to compute the full gradients, even though a full pass can be relaxed to a very large mini-batch for large data sets. 
For smooth but not strongly convex objectives, SVRG was shown to achieve the asymptotic convergence rate $\mathcal{O}(1/T)$, compared to $\mathcal{O}(1/\sqrt{T})$ of SGD. Many other control variates are used in practice \cite{ paisley2012variational,tucker2017rebar,oates2017control}. We present another popular type of a control variate, the score function control variate, in Section \ref{sec:BBVI}.

\vspace{10pt}
\subparagraph{Non-uniform Sampling.}
Instead of subsampling data points with equal probability,  
non-uniform sampling can be used to select mini-batches with a lower gradient variance.
Several authors suggested variants of importance sampling in the context of mini-batch selection
\cite{csiba2016importance, perekrestenko2017faster, zhao2015stochastic, gopalan2012scalable}. Although effective, these methods are not always practical, as the computational complexity of the sampling mechanism relates to the dimensionality of model parameters \cite{fu2017CPSGMCMC}. 
Alternative methods aim at de-correlating similar points and sampling diversified mini-batches. These methods include stratified sampling  \cite{zhao2014accelerating}, 
where one samples data from pre-defined subgroups based on meta-data or labels,  clustering-based sampling  \cite{fu2017CPSGMCMC}, which 
amounts to clustering the data using k-means and then sampling data from every cluster with adjusted probabilities, and diversified mini-batch sampling \cite{zhang2017stochastic,zhang2018active} using repulsive point processes to suppress the probability of data points with similar features in the same mini-batch. All of these methods have been shown to reduce variance and can also be used for learning on imbalanced data. 

\vspace{10pt}
\subparagraph{Other Methods.}
A number of alternative methods have been developed that contribute to variance reduction for SVI. A popular approach relies on Rao-Blackwellization, which is used in \cite{ranganath2014black}. The Rao-Blackwellization theorem (see Appendix \ref{sec:raoBlackwell}) generally states that a conditional estimation has lower variance if there exists a valid statistic that it can be conditioned on. Inspired by Rao-Blackwellization, the local expectation gradients method \cite{titsias2015local} has been proposed. This method splits the computation of the gradient of the ELBO into a Monte Carlo estimation and an exact expectation so that the contribution of each latent dimension to the gradient variance is optimally taken into account.
Another related approach has been developed for SVI, which averages expected sufficient statistics over a sliding window of mini-batches to obtain a natural gradient with smaller mean squared error \cite{mandt2014smoothed}.

\subsection{Collapsed, Sparse, and Distributed VI} 
\label{sec:modelspecific}

In contrast to using stochastic optimization 
for faster convergence, 
this section presents methods that leverage the structure of certain models to achieve the same goal. 
In particular, we focus on \emph{collapsed}, \emph{sparse}, \emph{parallel}, and \emph{distributed} inference.

\paragraph{Collapsed Inference}
Collapsed variational inference (CVI) relies on the idea of analytically integrating out certain model parameters \cite{teh2006collapsed,kurihara2007collapsed,sung2008latent,titsias2011variational,lazaro2012overlapping,hensman2012fast,king2006fast}. Due to the reduced number of parameters to be estimated, inference is typically faster. Collapsed inference is commonly constrained in the traditional conjugate exponential families, where 
the ELBO assumes an analytical form during marginalization. For these models, one can either marginalize out these latent variables before the ELBO is derived, or eliminate them afterwards \cite{king2006fast,hensman2012fast}.

Several authors have proposed CVI for topic models \cite{teh2006collapsed,kurihara2007collapsed} where one can either collapse the topic proportions \cite{teh2006collapsed} or the topic assignments \cite{hensman2012fast}.
In addition to these model specific derivations, \cite{hensman2012fast} unifies existing model-specific CVI approaches and presents a general collapsed inference method for models in the conjugate exponential family class. 
 
The computational benefit of CVI depends strongly on the statistics of the collapsed variables. 
Additionally, collapsing latent random variables can make other inference techniques tractable.  For models such as topic models, we can collapse the discrete variables and only infer the continuous ones. This allows the usage of inference networks  (Section \ref{sec:DAIforDL}) \cite{miao2016neural,srivastava2017autoencoding}. 

More generally, CVI does not solve all problems.
On the one side, integrating out certain model variables makes the ELBO tighter, since the marginal likelihood does not have to get lower-bounded in these variables. On the other hand, besides mathematical challenges, marginalizing variables can introduce additional dependencies between variables. For example, collapsing the global variables in Latent Dirichlet Allocation introduces non-local dependencies between the assignment variables, making distributed inference harder.

\paragraph{Sparse Inference}
Sparse inference introduces additional low-rank approximations into the variational approach, enabling more scalable inference 
\cite{snelson2006sparse,titsias2009variational,hensman2013gaussian}.  
Sparse inference can be either interpreted as a modeling choice or as an inference scheme \cite{bui2016unifying}.

Sparse inference methods are often encountered in the Gaussian Process (GPs) literature. 
The computational cost of learning GPs is $\mathcal{O}(M^3)$, where $M$ is the number of data points. This cost is caused by the inversion of the kernel matrix $K_{MM}$ of size $M\times M$, which hinders the application of GPs to big data sets. 
The idea of sparse inference in GPs \cite{snelson2006sparse} is to introduce $T$ \emph{inducing points}. 
Inducing points can be interpreted as pseudo-inputs that reflect the original data, but yield a more sparse representation since $T \ll M$. 
With inducing points, only a $T \times T$ sized matrix needs to be inverted, and consequently the computational complexity of this method is $\mathcal{O}(MT^2)$. 
\cite{titsias2009variational} collapses the distribution of inducing points, and  \cite{hensman2013gaussian} further extends this work to a stochastic version with a computational complexity of $\mathcal{O}(T^3)$. Additionally, sparse inducing points make inference in Deep GPs tractable \cite{damianou2012deep}.

\paragraph{Parallel and Distributed Inference}
Variational inference can be adjusted to distributed computing scenarios, where subsets of the data or parameters are distributed among several machines.
\cite{nallapati2007parallelized,zhai2012mr,gal2014distributed,neiswanger2015embarrassingly,broderick2013streaming}. Distributed inference schemes are often required in large scale scenarios, where data and computations are distributed across several machines. Independent latent variable models are trivially parallelizable. However, model specific designs such as reparametrizations might be required to enable efficient distributed inference \cite{gal2014distributed}. 
Current computing resources make VI applicable to web-scale data analysis \cite{zhai2012mr}.

\section{Generic VI: Beyond the conjugate exponential family}
\label{sec:generalizability}

In this section, we review techniques which aim at making VI more generic. 
This includes making VI applicable to a broader class of models, and also to make VI more automatic, eliminating the need for model-specific calculations.
This makes VI more accessible and easier to use.

Variational inference was originally limited to conditionally conjugate models, for which the ELBO could be computed analytically before it was optimized~\cite{hoffman13,zhang2016structured}. In this section, we introduce methods that relax this requirement and simplify inference. Central to this section are stochastic gradient estimators of the ELBO that can be computed for a broader class of models.

We start with the Laplace approximation in Section  \ref{sec:historygen} and illustrate its limitations. We will then introduce black box variational inference (BBVI) which removes the need for analytic solutions. We discuss BBVI methods that rely on the  REINFORCE or score function gradient in Section  \ref{sec:BBVI} and 
  a different form of BBVI, which uses reparameterization gradients, in Section  \ref{sec:BBVI_rep}. Other approaches for non-conjugate VI 
are discussed in Section  \ref{sec:otherGernicVI}. 

\subsection{Laplace's Method and Its Limitations}
\label{sec:historygen}

While not being the main focus of this survey, we briefly review the Laplace approximation as an alternative to non-conjugate inference.
The Laplace (or Gaussian) approximation approximates the posterior by a Gaussian distribution \cite{laplace1986memoir}. To this end, one seeks the maximum of the posterior and computes the inverse of its Hessian. These two entities are taken as the mean and covariance of the Gaussian posterior approximation. To make this approach feasible, the log posterior needs to be twice-differentiable. According to the Bernstein von Mises theorem (a.k.a.~ Bayesian central limit theorem) \cite{le2012asymptotic}, the posterior approaches a Gaussian asymptotically in the limit of large data, and the Laplace approximation becomes exact (provided that the model is under-parameterized). The approach can be applied to approximate the maximum a posteriori (MAP) mean and covariance, predictive densities, and marginal posterior densities \cite{tierney1986accurate}. The Laplace method has also been extended to more complex models such as belief networks with continuous variables \cite{azevedo1994laplace}. 

This approximation suffers mainly from being purely local and depending only on the curvature of the posterior around the optimum; KL minimization typically approximates the posterior covariance more accurately.
Additionally, the Laplace approximation is limited to the Gaussian variational family and does not apply to discrete variables \cite{wang2013variational}. Computationally, the method requires the inversion of a potentially large Hessian, which can be costly in high dimensions. This makes this approach intractable in setups with a large number of parameters.

\subsection{REINFORCE Gradients}
\label{sec:BBVI}
In classical VI, the ELBO is first derived analytically, and then optimized. This procedure is usually restricted to models in the conditionally conjugate exponential family \cite{hoffman13}. For many models, including
Bayesian deep learning architectures or complex hierarchical models, the ELBO contains intractable expectations with no known or simple analytical solution. Even if an analytic solution is available, the analytical derivation of the ELBO often requires time and mathematical expertise. In contrast, BBVI proposes a generic inference algorithm for which only the generative process of the data has to be specified. The main idea is to represent the gradient as an expectation, and to use Monte Carlo techniques to estimate this expectation.

As discussed in Section  \ref{sec:ingredients}, in general VI aims at maximizing the ELBO, which is equivalent to minimizing the KL divergence between the variational posterior and  target distribution.  
To maximize the ELBO, one needs to follow the gradient or stochastic gradient of the variational parameters.
 The key insight of BBVI is that one can obtain an unbiased gradient estimator by sampling from the variational distribution without having to compute the ELBO analytically\cite{paisley2012variational,ranganath2014black}. 

For a broad class of models, the gradient of the ELBO can be expressed as an expectation with respect to the variational distribution~\cite{ranganath2014black}:
\begin{equation}
\nabla_{\bmlambda} \mathcal{L}= \E_q [\nabla_{\bmlambda} \log q( \bmz | \bmlambda) (  \log p(\bmx, \bmz) - \log q(\bmz|\bmlambda)  ) ].
\label{eq:BBVI}
\end{equation} 
The full gradient $\nabla_{\bmlambda} \mathcal{L}$, involving the expectation over $q$, can now be approximated by a stochastic gradient estimator $\nabla_{\bmlambda} \hat{\mathcal{L}_s}$ by sampling from $q$:
\begin{equation}
\nabla_{\bmlambda} \hat{\mathcal{L}_s} =  \frac{1}{K} \sum _{k=1} ^ K  \nabla_{\bmlambda} \log q(\z_k | \bmlambda) (  \log p(\bmx, z_k) - \log q(z_k|\bmlambda)  ), \label{eq:BBVIgradient}
\end{equation} 
where  $z_k \sim q(\bmz | \bmlambda )$. Thus, BBVI provides black box gradient estimators for VI. Moreover, it only requires the practitioner to provide the joint distribution of observations and latent variables without the need to derive the gradient of the ELBO explicitly. The quantity $\nabla_{\bmlambda} \log q(z_k | \bmlambda)$ is also known as the score function and is part of the REINFORCE algorithm \cite{williams1992simple}. 

The derivation of Eq. \ref{eq:BBVI} requires the  log derivative trick  which can be applied to any bound. While the ELBO in combination with the KL results in  Eq. \ref{eq:BBVI}, other divergence measures lead to additional terms in the REINFORCE gradients (B.3 in \cite{li2018Approximate}).

A direct implementation of stochastic gradient ascent based on Eq. \ref{eq:BBVIgradient} suffers from high variances of the estimated gradients. Much of the success of BBVI can be attributed to variance reduction through Rao-Blackwellization and  control variates  \cite{ranganath2014black}.
As one of the most important advancements of modern approximate inference, BBVI as been extended and made amortized inference feasible, see Section \ref{sec:amortizedVI}.

\paragraph{Variance Reduction for BBVI}
\label{sec:variancereduction}

BBVI requires a different set of techniques for variance reduction than those reviewed for SVI in Section  \ref{sec:FactConvergence}. 
In contrast to SVI where the noise resulted from subsampling from a finite set of data points, the BBVI noise originates from random variables with possibly infinite support. In this case, techniques such as SVRG are not applicable, since the full gradient is not a sum over finitely many terms and cannot be kept in memory. Hence, BBVI involves a different set of control variates and other methods which shall briefly be reviewed here.

The arguably most important control variate in BBVI is the score function control variate \cite{ranganath2014black}, where one subtracts a Monte Carlo expectation of the score function from the gradient estimator:
\begin{equation}
\nabla_{\bmlambda} \hat{\mathcal{L}}_{control} = \nabla_{\bmlambda} \hat{\mathcal{L}} -   \frac{w}{K} \sum _{k=1} ^ K \nabla_{\bmlambda}\log q(z_k|\bmlambda)
\label{eq:BBVIcontrol}
\end{equation} 

As required, the score function control variate has expectation zero under the variational distribution. The weight $w$ is selected such that it minimizes the variance of the gradient. 

While the original BBVI paper introduces both Rao-Blackwellization and control variates, \cite{titsias2015local} points out that good choices for control variates  might be model-dependent. They further elaborate on local expectation gradients, which take only the Markov blanket of each variable into account. A different approach is presented by \cite{ruiz2016overdispersed}, which introduces overdispersed importance sampling. By sampling from a proposal distribution that belongs to an overdispersed exponential family and that places high mass on the tails of the variational distribution, the variance of the gradient is reduced.

\subsection{Reparameterization Gradient VI}
\label{sec:BBVI_rep}

An alternative to the REINFORCE gradients introduced in Section~\ref{sec:BBVI} are the so-called reparameterization gradients. These gradients are obtained by representing the variational distribution as a deterministic parametric transformation of a noise distribution.
Empirically, reparameterization gradients are often found to have lower variance than REINFORCE gradients.

\paragraph{Reparameterization Gradients}
\label{sec:reparameterization}

The reparameterization trick allows to estimate the gradient of the ELBO by Monte Carlo samples by
representing random variables as deterministic functions of noise distributions. This gives low-variance stochastic gradients for a large class of models without having to compute analytic expectations.

In more detail, the trick states that a random variable $z$ with a distribution $q(z ;\bmlambda )$ can be expressed as a transformation of a random variable $\epsilon \sim r(\epsilon)$ that comes from a noise distribution, such as uniform or Gaussian.
For example, if $z \sim \mathcal{N}(z;\mu, \sigma^2)$, then 
$z = \mu + \sigma \epsilon$ where $\epsilon \sim \mathcal{N}(\epsilon;0,1)$ \cite{kingma2013auto,rezende2014stochastic}. 

More generally, the random variable $z$ is given by a parameterized, deterministic function of random noise, $z = g(\epsilon, \bmlambda), \, \epsilon \sim r(\epsilon)$. Importantly, the noise distribution $p(\epsilon)$ is considered independent of the parameters of $q(z ;\bmlambda )$, and therefore $q(z ;\bmlambda )$ and $g(\epsilon, \bmlambda)$ share the same parameters $\bmlambda$. This allows to compute any expectation over $z$ as an expectation over $\epsilon$  by the theory behind the change of variables in integrals. 
We can now build a stochastic gradient estimator of the ELBO by pulling the gradient into the expectation, and approximating it by samples from the noise distribution:
\begin{align} 
\nabla_{\bmlambda} \hat{\mathcal{L}_{rep}} =  \frac{1}{K} \sum _{k=1} ^ K &   \nabla_{\bmlambda} \Big( \log p(\x_i, g(\epsilon_k,  \bmlambda)) - \nonumber \\
&\log q(g(\epsilon_k,  \bmlambda)|\bmlambda)\Big), ~~\epsilon_k \sim r(\epsilon).\label{eq:reparagradient}
\end{align}
Often, the entropy term can be computed analytically, which can lead to a lower gradient variance \cite{kingma2013auto}. 

Note that the gradient of the log joint distribution enters the expectation. This is in contrast to the REINFORCE gradient, where the gradient of the variational distribution is taken (Eq.~\ref{eq:BBVIgradient}). 
The advantage of taking the gradient of the log joint is that this term is more informed about the direction of the maximum posterior mode. The lower variance of the reparameterization gradient may be attributed to this property.

 While the variance of this estimator (Eq. \ref{eq:reparagradient}) is often lower than the variance of the score function gradient (Eq. \ref{eq:BBVIgradient}), a theoretical analysis shows that this is not guaranteed, see Chapter 3 in
\cite{gal2016uncertainty}. \cite{roeder2017sticking} shows that the reparameterization gradient can be divided into a path derivative and the score function. Omitting the score function 
in the vicinity of the optimum can result in an unbiased gradient estimator with lower variance. 
Reparameterization gradients are also the key to variational autoencoders \cite{kingma2013auto, rezende2014stochastic}  which we discuss in detail in Section  \ref{sec:vae}.

The reparameterization trick does not trivially extend to many distributions, in particular to discrete ones. Even if a reparameterization function exists, it may not be differentiable. In order for the reparameterization trick to apply to discrete distributions, 
 the variational distributions require further approximations.
Several groups have addressed this problem.
In  \cite{jang2017categorical, maddison2017concrete}, the categorical distribution is approximated with the help of the Gumbel-Max trick and by replacing the argmax operation with a softmax operator. A temperature parameter controls the degree to which the softmax can approximate the categorical distribution. The closer it resembles a categorical distribution, the higher the variance of the gradient. The authors propose annealing strategies to improve convergence.  Similarly, a stick-breaking process is used in \cite{nalisnick2016stick} to approximate 
the Beta distribution with the Kumaraswamy distribution.

As many of these approaches rely on approximations of individual distributions, there is growing interest in more general methods that are applicable without specialized approximations. The generalized reparameterization gradient \cite{ruiz2016Generalized} achieves this by finding an invertible transformation between the noise and the latent variable of interest. The authors derive the gradient of the ELBO which decomposes the expected likelihood into the standard reparameterization gradient and a correction term. The correction term is only needed when the transformation  weakly depends on the variational parameters. A similar division is derived by \cite{naesseth2017reparameterization} which proposes an accept-reject sampling algorithm for reparameterization gradients that allows one to sample from expressive posteriors. While reparameterization gradients often demonstrate lower variance than the score function, the use of Monte Carlo estimates still suffers from the injected noise. The variance can be further reduced with control variates \cite{miller2017reducing, roeder2017sticking} or Quasi-Monte Carlo methods \cite{buchholz2018quasi}. 
 
\subsection{Other Generalizations}
\label{sec:otherGernicVI}

Finally, we survey a number of approaches that consider VI in non-conjugate models but do not follow the BBVI principle.
Since the ELBO for non-conjugate models contains intractable integrals, these integrals have to be approximated somehow, either using some form of Taylor approximations (including Laplace approximations), lower-bounding the ELBO further such that the resulting integrals are tractable, or using some form of Monte Carlo estimators.
Approximation methods which involve inner optimization routines \cite{blei2006correlated,wang2009simultaneous,zhang13c} often become prohibitively slow for practical inference tasks. In contrast, approaches based on additional lower bounds with closed-form updates
\cite{knowles2011non,wang2013variational,khan2015kullback} can be computationally more efficient.
Examples include extensions of the variational message passing algorithm \cite{winn2005variational} to non-conjugate models \cite{knowles2011non, wang2013variational}. Furthermore, \cite{salimans2013fixed}
proposed a technique based on stochastic linear regression to estimate the parameters of a fixed variational distribution based on Monte Carlo approximations of certain sufficient statistics.
Recently, \cite{khan2015kullback} proposed a hybrid approach, where inference is 
split into a conjugate and a non-conjugate part.

\section{Accurate VI: \\Beyond KL and Mean Field}
\label{sec:accuracy}

In this section, we present various methods that aim at improving the accuracy of standard VI. Previous sections dealt with making VI scalable and applicable to non-conjugate exponential family models. Most of the work in those areas, however, still addresses the standard setup of MFVI and employs the KL divergence as a measure of distance between distributions. Here we review recent developments that go beyond this setup, with the goal of avoiding poor local optima and increasing the accuracy of VI. Inference networks, normalizing flows, and related methods may also be considered as non-standard VI, but are discussed in Section \ref{sec:DAIforDL}.

We start by reviewing the origins of MFVI in statistical physics and describe its limitations (Section \ref{sec:AccuracyHistory}). We then discuss alternative divergence measures in Section \ref{sec:Divergence}. Structured variational approximations  beyond mean field are discussed in Section \ref{sec:structuredVI}, followed by alternative methods that do not fall into the previous two classes (Section \ref{sec:Acc_other}).
 
\subsection{Origins and Limitations of Mean Field VI}
\label{sec:AccuracyHistory}

Variational methods have a long tradition in statistical physics.
The mean field method was originally applied to model spin glasses, which are certain types of disordered magnets where the magnetic spins of the atoms are not aligned in a regular pattern~\cite{opper2001advanced}. A simple example for such a spin glass model is the Ising model, a model of binary variables on a lattice with pairwise couplings. 
To estimate the resulting statistical distribution of spin states, a simpler, factorized distribution is used as a proxy. This is done 
with the goal of approximating the marginal probabilities of the spins pointing up or down (also called 'magnetization') as well as possible, while ignoring all correlations between the spins. 
The many interactions of a given spin with its neighbors are replaced by a single interaction between a spin and the effective magnetic field (a.k.a. \emph{mean field}) of all other spins. This explains the name origin.
 Physicists typically denote the negative log posterior as an energy or Hamiltonian function. This language has been adopted by the machine learning community for approximate inference in both directed and undirected models, summarized in Appendix \ref{sec:Physics_Notations} for the reader's reference. 

Mean field methods were first adopted in neural networks by Anderson and Peterson in 1987 \cite{peterson1987mean}, and later gained popularity in the machine learning community \cite{saul1996mean,jordan1999introduction,opper2001advanced}. The main limitation of mean field approximations is that they 
explicitly ignore correlations between different variables e.g., between the spins in the Ising model. Furthermore, \cite{wainwright2008graphical} showed that 
the more possible dependencies are broken by the variational distribution, 
the more non-convex the optimization problem becomes.
Conversely, if the variational distribution contains more structure, certain local optima do not exist.
A number of initiatives to improve mean field VI have been proposed by the physics community and further developed by the machine learning community \cite{thouless1977solution,plefka1982convergence,opper2001advanced}. 

An early example of going beyond the mean field theory in a spin glass system is the Thouless-Anderson-Palmer (TAP) equation approach \cite{thouless1977solution}, which introduces perturbative corrections to the variational free energy.
A related idea relies on power expansions \cite{plefka1982convergence}, which has been extended and applied to machine learning models by various authors \cite{kappen2001second,opper2001tractable,opper2013perturbative,raymond2014expectation,tanaka1998estimation}. Additionally, information geometry provides an insight into the relation between MFVI and TAP equations \cite{tanaka1999theory,tanaka2000information}. \cite{zhang2017perturbative} further connects TAP equations with divergence measures. 
We refer the readers to \cite{opper2001advanced} for more information. Next, we review the recent advances beyond MFVI based on divergence measures other than the KL divergence.

\subsection{VI with Alternative Divergences}
\label{sec:Divergence}

The KL divergence often provides a computationally convenient method to measure the distance between two distributions. It leads to analytically tractable expectations for certain model classes. However, traditional Kullback-Leibler variational inference (KLVI) suffers from problems such as underestimating posterior variances \cite{minka05divergence}. In other cases, it is unable to break symmetry when multiple modes are close \cite{opper2015perturbation}, and is  a comparably loose bound \cite{zhang2017perturbative}. Due to these shortcomings, a number of other divergence measures have been proposed which we survey here. 
 
New divergence measures beyond the KL divergence do not only play a role in VI, but also in related approximate inference methods such as EP. Some recent extensions of EP
\cite{minka2004power, wainwright2005new, li2015stochastic,zhe2016online} 
can be viewed as classical EP with alternative divergence measures \cite{minka05divergence}.
While these methods are sophisticated, a practitioner will find them difficult to use due to complex derivations and limited scalability.  Recent developments of VI focus mainly on a unified framework in a black box fashion to allow for scalability and accessibility. BBVI rendered the application of other divergence measures, such as the $\chi$ divergence \cite{dieng2016chi}, possible while maintaining the efficiency and simplicity of the method.

In this section, we introduce relevant divergence measures and show how to use them in the context of VI. 
The KL divergence, as discussed in Section \ref{sec:KLDivergence}, is a special form of the \talph-divergence, while the \talph-divergence is a special form of the $f$-divergence. All above divergences can be written in the form of the Stein discrepancy.

\paragraph{\talph-divergence}
The \talph-divergence is a family of divergence measures with interesting properties from an information geometrical and computational perspective \cite{amari1985differential, amari2009divergence}. Both the KL divergence and the Hellinger distance are special cases of the \talph-divergence.  

Different formulations of the  \talph-divergence exist\cite{zhu1995information, amari2009divergence}, and various VI methods use different definitions \cite{li2016black, minka05divergence}.  We focus on Renyi's formulation, 
\begin{equation}
D_\alpha ^R (p || q) = \frac{1}{\alpha - 1} \log \int p(x) ^ \alpha q(x) ^ {1 - \alpha} d x,
\end{equation}
where  $\alpha > 0, \alpha \neq 1$. 
With this definition of \talph-divergences, a smaller $\alpha$ leads to more mass-covering effects, while a larger $\alpha$ results in zero-forcing effects, meaning that the variational distribution avoids areas of low posterior probability. For $\alpha \rightarrow 1$,  we recover standard VI (involving the KL divergence).

\talph-divergences have recently been used in variational inference \cite{li2016black,li2016renyi}.
Similar as in the derivation of the ELBO in Eq.~\ref{eq:ELBO_KL}, the $\alpha$-divergence implies a bound on the marginal likelihood:
\begin{align}
\mathcal{L}_{\alpha} &= \log p(\bm{x}) - D_\alpha ^R (q(\bm{z}) ||p(\bm{z} | \bm{x})) \nonumber \\
&= \frac{1}{\alpha - 1 } \log \E _q \left[ \left( \frac{p(\bm{z} , \bm{x})}{q(\bm{z})} \right) ^ {1-\alpha} \right].
\label{eq:alpha_bound}
\end{align}
For $\alpha \geq 0, \alpha \neq 1$, $\mathcal{L}_\alpha$ is a lower bound on the log marginal likelihood. Interestingly, Eq.~\ref{eq:alpha_bound} also admits negative values of $\alpha$, in which case it becomes an \emph{upper} bound. Note that in this case, $D_\alpha ^R$ is not a divergence. 
Among various possible definitions of the $\alpha$-divergence, only Renyi's formulation leads to a bound (Eq. \ref{eq:alpha_bound}) in which the marginal likelihood $p(x)$ cancels.

\paragraph{$f$-Divergence and Generalized VI}
$\alpha$-divergences are a subset of the more general family of $f$-divergences \cite{csiszar1964informationstheoretische,ali1966general}, which take the form:
\begin{align*}
D_f(p || q) = \int q(x) f \left( \frac{p(x)}{q(x)} \right) dx.
\end{align*}
$f$ is a convex function with $f(1)=0$. For example, the KL divergence $KL(p || q)$ is represented by the $f$-divergence with $f(r) = r \log(r)$, 
and the Pearson $\chi ^2$ distance is an $f$-divergence with $f(r) = (r-1)^2$. 

In general, only specific choices of $f$ result in a bound that does only trivially depend on the marginal likelihood, and which is therefore useful for VI.

\cite{zhang2017perturbative} lower-bounded the marginal likelihood using Jensen's inequality:
\begin{align}
\label{eq:biased-bound}
p(\bm{x}) \geq \ff (p(\bm{x})) \geq  \E_{q(\bm{z})} \left[\ff \left(\frac{p(\bmx,\bmz)}{q(\bmz)}\right)\right] \equiv {\cal L}_{\ff}.
\end{align}
Above, $\ff$ is an arbitrary concave function with $\ff(x)< x$. This formulation recovers the true marginal likelihood for $\ff = id$, the standard ELBO for ${\ff} = \log$, and $\alpha$-VI for ${\ff}(x) \propto  x^{(1-\alpha)}$. 
For $V \equiv \log q(\bmz) - \log p(\bmx,\bmz)$, the authors propose the following function:
\begin{align*} 
&{\ff}^{(V_0)}(e^{-V})\\
&=e^{-V_0}\Big(1 + (V_0-V) + \frac{1}{2} (V_0-V)^2 + \frac{1}{6}(V_0-V)^3 \Big).
\end{align*}
Above, $V_0$ is a free parameter that can be optimized, and which absorbs the bound's dependence on the marginal likelihood. 
The authors show that the terms up to linear order in $V$ correspond to the KL divergence, whereas higher order polynomials are correction terms which make the bound tighter. This connects to earlier work on TAP equations \cite{thouless1977solution,plefka1982convergence} (see Section \ref{sec:AccuracyHistory}), which generally did not result in a bound.

\paragraph{Stein Discrepancy and VI}
Stein's method \cite{stein1972bound} was first proposed as an error bound to measure how well an approximate distribution fits a distribution of interest. The Stein discrepancy has been adapted to modern VI \cite{liu2016kernelized,liu2016stein,liu2017stein,han2017stein}.
Here, we introduce the Stein discrepancy and two VI methods that use it: Stein Variational Gradient Descent (SVDG) \cite{liu2016stein} and operator VI \cite{ranganath2016Operator}. These two methods share the same objective but are optimized in different manners.
 
The Stein discrepancy is an integral probability metric \cite{muller1997integral,sriperumbudur2009integral,mescheder2017numerics}.
In particular, \cite{ranganath2016Operator,liu2016stein} used the Stein discrepancy as a divergence measure: 
\begin{equation}
D_{\text{stein}} (p, q) = sup_{f \in \mathcal{F}} | \E_{q(\bmz)}[f(\bmz)] - \E _ {p(\bmz|\bmx)}[f(\bmz)]|^2.
\label{eq:D_stein_def}
\end{equation}
$\mathcal{F}$ indicates a set of smooth, real-valued functions. When $q(\bmz)$ and $p(\bmz|\bmx)$ are identical, the divergence is zero. More generally, the more similar $p$ and $q$ are, the smaller is the discrepancy.

The second term in Eq. \ref{eq:D_stein_def} involves an expectation under the intractable posterior. 
Therefore, the Stein discrepancy can only be used in VI for classes of functions $\mathcal{F}$ for which the second term is equal to zero. We can find a suitable class with this property as follows. We define $f$ by applying a differential operator $\mathcal{A}$ on another function  $\phi$, where $\phi$ is only restricted to be smooth:
\begin{equation*}
f(\bmz) = \mathcal{A}_p \phi(\bmz),
\end{equation*}
where $\bmz \sim p(\bmz)$. The operator $\mathcal{A}$ is constructed in such a way that the second expectation in Eq.~\ref{eq:D_stein_def} is zero for arbitrary $\phi$;
all operators with this property are valid operators  \cite{ranganath2016Operator}. A popular operator that fulfills this requirement is the Stein operator: 
\begin{equation*}
\mathcal{A}_p \phi(\bmz) = \phi(\bmz)\nabla_{\bmz} \log p(\bmz, \bmx) + \nabla _{\bmz} \phi(\bmz).
\end{equation*}
 Both operator VI \cite{ranganath2016Operator} and  SVGD \cite{liu2016stein} use the Stein discrepancy with the Stein operator to construct the variational objective.
The main difference between these two methods lies in the optimization of the variational objective using the Stein discrepancy. Operator VI \cite{ranganath2016Operator} uses a minimax (GAN-style) formulation and BBVI to optimize the variational objective directly; while Stein Variational Gradient Descent (SVGD) \cite{liu2016stein} uses a kernelized Stein discrepancy. With a particular choice of the kernel and $q$, it can be shown that SVGD determines the optimal perturbation in the direction of the steepest gradient of the KL divergence \cite{liu2016stein}. 
SVGD leads to a scheme where samples in the latent space are sequentially transformed to approximate the posterior. As such, the method is reminiscent of, though formally distinct from, a  normalizing flow approach \cite{rezende2015variational} (see Section \ref{sec:advanceVAE}).

\subsection{Structured Variational Inference}
\label{sec:structuredVI}

MFVI assumes a fully-factorized variational distribution; as such, it is unable to capture posterior correlations. Fully factorized variational models have limited accuracy, especially when the latent variables are highly dependent such as in models with hierarchical structure.
This section examines variational distributions which are not fully factorized, but contain dependencies between the latent variables. These structured variational distributions are more expressive, but often come at higher computational costs. 

Allowing a structured variational distribution to capture dependencies between latent variables is a modeling choice; different dependencies may be more or less relevant and depend on the model under consideration.
For example, structured variational inference for
LDA \cite{hoffman2014structured} shows that maintaining global structure is vital, while structured variational
inference for the Beta Bernoulli Process \cite{shah2015empirical} shows that maintaining local structure is more important.
As follows, we review structured inference for hierarchical models, and discuss VI for time series.

\paragraph{Hierarchical VI} 
For many models, the variational approximation can be made more expressive by maintaining dependencies between latent variables, but these dependencies make it harder to estimate the gradient of the variational bound.
Hierarchical variational models (HVM)\cite{ranganath2016hierarchical} are a black box VI framework for structured variational distributions which applies to a broad class of models. In order to capture dependencies between latent variables, one starts with a mean-field variational distribution $\prod _i q(z_i;\lambda_i)$, but instead of estimating the variational parameters $\bm{\lambda}$, one places a prior $q(\bm{ \lambda}; \bm{\theta} )$ over them and marginalizes them out:
\begin{equation}
q(\bm{z}; \bm{\theta}) = \int \left( \prod _i q(z_i; \lambda_i)\right) q(\bm{ \lambda} ; \bm{\theta} ) d \bm{\lambda}.
\label{eq:HMF}
\end{equation}
The new variational distribution  $q(\bm{z}; \bm{\theta} )$ thus captures dependencies through the marginalization procedure. Sampling from this distribution is also possible by simulating the hierarchical process. 
The resulting ELBO can be made tractable by further lower-bounding the resulting entropy and sampling from the hierarchical model. 
Notably, this approach is used in the development of the variational Gaussian Process (VGP) \cite{tran2016variational}, a particular HVM. The VGP applies a Gaussian Process to generate variational estimates, thus forming a  Bayesian non-parametric prior. Since GPs can model a rich class of  functions, the VGP is able to confidently approximate diverse posterior distributions \cite{tran2016variational}.  


{Another method that established dependencies between latent variables is copula VI \cite{tran2015copula,han2016variational}.
Instead of using a fully factorized variational distribution, copula VI assumes the variational family form:
\begin{equation}
q(\bm{z}) =  \left( \prod _i q(z_i; \lambda_i)\right) c \left( Q(z_1), ... , Q(z_N) \right),
\end{equation}
where $c$ is the copula distribution, which is a joint distribution over the marginal cumulative distribution functions $Q(z_1), ... , Q(z_N)$. This copula distribution restores the dependencies among the latent variables. 
}

\paragraph{VI for Time Series} 
One of the most important model classes in need of structured variational approximations are time series models.
Significant examples include Hidden Markov Models (HMM)\cite{eddy1996hidden} and Dynamic Topic Models (DTM) \cite{Blei06D}.
These models have strong dependencies between time steps, leading traditional fully factorized MFVI  to produce unsatisfying results. 
When using VI for time series, one typically employs a structured variational distribution that explicitly captures dependencies between time points, while remaining fully-factorized in the remaining variables \cite{Blei06D,johnson2014stochastic, foti2014stochastic,bamler2017structured}. This commonly requires model specific approximations. \cite{johnson2014stochastic,foti2014stochastic} derive SVI for popular time series models including HMMs, hidden semi-Markov models (HSMM), and hierarchical Dirichlet process-HMMs. Moreover, \cite{johnson2014stochastic} derive an accelerated SVI for HSMMs. 
\cite{bamler2017structured, bamler2017dynamic} derive a structured BBVI algorithm for non-conjugate latent diffusion models. 
 
\subsection{Other Non-Standard VI Methods}
\label{sec:Acc_other}

In this section, we cover a number of 
miscellaneous approaches which fall under the broad umbrella of improving the accuracy of VI, but would not be categorized as alternative divergence measures or structured models.

\paragraph{VI With Mixture Distributions}

Mixture distributions form a class of very flexible distributions, and have been used in VI since the 1990s \cite{jaakkola1998improving,jordan1999introduction}.  Due to their flexibility as well as computational difficulties, advancing VI for mixture models has been of continuous interest \cite{gershman2012nonparametric,salimans2013fixed, guo2016boosting,miller2016variational,arenz2018efficient}.  To fit a mixture model, we can make use of auxiliary bounds \cite{ranganath2016hierarchical}, a fixed point update \cite{salimans2013fixed}, or enforce additional assumptions such as using uniform weights \cite{gershman2012nonparametric}. Inspired by boosting methods, recently proposed methods fit mixture components in a successive manner \cite{guo2016boosting,miller2016variational}. Here, Boosting VI and variational boosting \cite{guo2016boosting,miller2016variational} refine the approximate posterior iteratively by adding one component at a time while keeping previously fitted components fixed. In a different approach, \cite{arenz2018efficient} utilizes stochastic policy search methods found in the Reinforcement Learning literature for fitting Gaussian mixture models.

\paragraph{VI by Stochastic Gradient Descent}
\label{sec:VIandSampling}

Stochastic gradient descent on the negative log posterior of a probabilistic model can, under certain circumstances, be seen as an implicit VI algorithm. Here we consider SGD with constant learning rates (constant SGD) \cite{mandt2016variational,mandt2017stochastic}, and early stopping \cite{duvenaud2016early}. 

Constant SGD can be viewed as a Markov chain that converges to a stationary distribution; as such, it resembles Langevin dynamics \cite{welling2011bayesian}. The variance of the stationary distribution is controlled by the learning rate. \cite{mandt2016variational} shows that the learning rate can be tuned to minimize the KL divergence between the resulting stationary distribution and the Bayesian posterior. Additionally,  \cite{mandt2016variational} derive formulas for this optimal learning rate which resemble AdaGrad \cite{duchi2011adaptive} and its relatives. A generalization of SGD that includes momentum and iterative averaging is presented in \cite{mandt2017stochastic}.  In contrast, \cite{duvenaud2016early} interprets SGD as a non-parametric  VI scheme. The paper proposes a way to track entropy changes in the implicit variational objective based on estimates of the Hessian. As such, the authors consider sampling from distributions that are not stationary.

\paragraph{Robustness to Outliers and Local Optima}
\label{sec:robust}
Since the ELBO is a non-convex objective, VI benefits from advanced optimization algorithms that help to escape from poor local optima.
Variational tempering \cite{mandt2016variationalTem} adapts deterministic annealing \cite{rose1990deterministic, neal1993probabilistic}to VI, making the cooling schedule adaptive and data-dependent.
Temperature can be defined globally or locally, where local temperatures are specific to individual data points.
Data points with associated small likelihoods under the model (such as outliers) are automatically assigned a high temperature. This reduces their influence on the global variational parameters, making the inference algorithm more robust to local optima.
Variational tempering can also be interpreted as data re-weighting \cite{wang2016reweighted}, the weight being the inverse temperature. In this context, lower weights are assigned to outliers. 
Other means of making VI more robust
include the trust-region method \cite{theis15}, which uses the KL divergence to tune the learning progress and avoids poor local optima, and population VI \cite{kucukelbir2015population}, which  averages the variational posterior over bootstrapped data samples for more robust modeling performance.

\section{Amortized variational inference and deep learning}
\label{sec:DAIforDL}
Finally, we review amoritzed variational inference.
Consider the setup of Section~\ref{sec:scalability}, where each data point $x_i$ is goverend by its latent variable $z_i$ with variational parameter $\xi_i$.
Traditional VI 
makes it necessary to optimize 
a $\xi_i$ for each data point $x_i$, which is computationally expensive, in particular when this optimization is embedded a global parameter update loop. The basic idea behind amortized inference is to use a powerful predictor to predict the optimal $z_i$ based on the features of $x_i$, i.e., $z_i = f(x_i)$. This way, the local variational parameters are replaced by a function of the data whose parameters are shared across all data points, i.e. inference is \emph{amortized}.

We detail the main ideas behind this approach in Section \ref{sec:amortizedVI}, and show how it is applied in form of variational autoencoders in Sections \ref{sec:vae} and \ref{sec:advanceVAE}.

\subsection{Amortized Variational Inference}
\label{sec:amortizedVI}

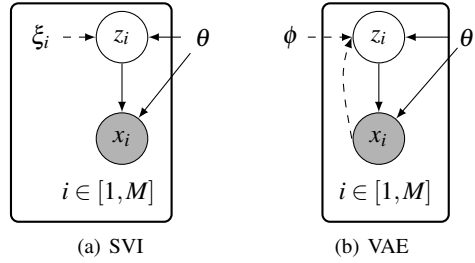
\begin{figure}[t!]
\centering
\subfigure[SVI]{
\pgfdeclarelayer{background}
\pgfdeclarelayer{foreground}
\pgfsetlayers{background,main,foreground}

\begin{tikzpicture}[scale=0.9]
\tikzstyle{surround} = [thick,draw=black,rounded corners=1mm]
\tikzstyle{scalarnode} = [circle, draw, fill=white!11,  
    text width=1.2em, text badly centered, inner sep=2.5pt]
\tikzstyle{scalarnodenoline} = [  fill=white!11, 
    text width=1.2em, text badly centered, inner sep=2.5pt]
\tikzstyle{arrowline} = [draw,color=black, -latex]
\tikzstyle{dashedarrowcurve} = [draw,color=black, dashed, out=100,in=250, -latex]
\tikzstyle{dashedarrowline} = [draw,color=black, dashed,  -latex]

    \node [scalarnodenoline] at (1.2,0) (O)   {$\theta$};
    \node [scalarnodenoline] at (-1.2,0) (P)   {$\xi_i$};
    \node [scalarnode] at (0, 0) (Z) {$z_i$};
    \node [scalarnode, fill=black!30 ] at (0, -1.5) (X) {$x_i$};
    \path [arrowline]  (O) to (Z);
    \path [arrowline] (Z) to (X);
    \path[arrowline]  (O) to (X);
    \path[dashedarrowline]  (P) to (Z);
 
    \node [] at (-0.2, -2.3) (M) {$i \in [1,M]$};
    \node[surround, inner sep = .1cm] (f_N) [fit = (Z)(X)(P)(M) ] {};
\end{tikzpicture}
\label{fig:vae_svi}
}
\subfigure[VAE]{
\pgfdeclarelayer{background}
\pgfdeclarelayer{foreground}
\pgfsetlayers{background,main,foreground}

\begin{tikzpicture}[scale=0.9]
\tikzstyle{surround} = [thick,draw=black,rounded corners=1mm]
\tikzstyle{scalarnode} = [circle, draw, fill=white!11,  
    text width=1.2em, text badly centered, inner sep=2.5pt]
\tikzstyle{scalarnodenoline} = [  fill=white!11, 
    text width=1.2em, text badly centered, inner sep=0.5pt]
\tikzstyle{arrowline} = [draw,color=black, -latex]
\tikzstyle{dashedarrowcurve} = [draw,color=black, dashed, out=100,in=250, -latex]
\tikzstyle{dashedarrowline} = [draw,color=black, dashed,  -latex]

    \node [scalarnodenoline] at (1.3,0) (O)   {$\theta$};
    \node [scalarnodenoline] at (-1.3,0) (P)   {$\phi$};
    \node [scalarnode] at (0, 0) (Z) {$z_i$};
    \node [scalarnode, fill=black!30 ] at (0, -1.5) (X) {$x_i$};
    
    \path [arrowline]  (O) to (Z);
    \path [arrowline] (Z) to (X);
    \path[arrowline]  (O) to (X);
    \path[dashedarrowline]  (P) to (Z);
    \path[dashedarrowcurve]  (-0.4,-1.5) to (-0.4,0);
    \node [] at (0.1, -2.3) (M) { $i\in[1,M]$};
    \node[surround, inner sep = .1cm] (f_N) [fit = (Z)(X)(M) ] {};
\end{tikzpicture}

 
\label{fig:vae_vae}
}
\caption{ The graphical representation of  stochastic variational inference (a) and the variational autoencoder (b). Dashed lines indicate variational approximations. \vspace{-10pt}}
\label{fig:intro_EncoderDecoder}
\end{figure}

The term amortized inference refers to utilizing inferences from past computations to support future computations \cite{gershman2014amortized, dayan1995helmholtz}. For VI,  amortized inference usually refers to inference over
local variables. Instead of approximating separate variables for each data point, as shown in Figure \ref{fig:vae_svi}, amortized VI assumes that the local variational parameters can be predicted by a parameterized function of the data. Thus, once this function is estimated, the latent variables can be acquired by passing new data points through the function, as shown in Figure \ref{fig:vae_vae}. Deep neural networks used in this context are also called \emph{inference networks}. Amortized VI with inference networks thus combines probabilistic modeling with the representational power of deep learning.
 
As an example, amortized inference has been applied to Deep Gaussian Processes (DGPs) \cite{damianou2012deep}.
Inference in these models is intractable, which is why the authors apply MFVI with inducing points (see Section~\ref{sec:modelspecific}) \cite{damianou2012deep}. 
Instead of estimating these latent variables separately, however, \cite{daiVAEDGP16} proposes to estimate these latent variables as functions of inference networks, allowing DGPs to scale to bigger datasets and speeding up convergence. Amortization can be also made iterative by feeding back the amortization error into the inference model \cite{pmlr-v80-marino18a}.


\subsection{Variational Autoencoders}
\label{sec:vae}

Amortized VI has become a popular tool for inference in deep latent Gaussian models (DLGM). This leads to the concept of variational autoencoders (VAEs), which have been proposed independently by two groups  \cite{kingma2013auto,rezende2014stochastic}, and which are discussed in detail below. VAEs apply more generally than to DLGMs, but for simplicity we will restrict our discussion to this model class.

\paragraph{The Generative Model}

In this paragraph we introduce the class of deep latent Gaussian models. The corresponding graphical model is depicted in Figure \ref{fig:vae_vae}.
The model employs a multivariate normal prior from which we draw  a latent variable $z$, 
$$p(z) = \mathcal{N}(0, \mathbb{I}).$$ 
More generally, this could be an arbitrary prior $p_\theta(z)$ that depends on additional parameters $\theta$. The likelihood of the model is: 
$$p_\theta(\bmx|\bmz)  =  \prod_{i=1}^N\mathcal{N}(x_i; \mu(z_i), \sigma^2(z_i)\mathbb{I}).
$$
Most importantly, the likelihood depends on $\bmz$ through two non-linear functions $\mu(\cdot)$ and $\sigma(\cdot)$. These are typically neural networks, which take the latent variables as an input and transform them in a non-linear way. The data are then drawn from a normal distribution centered around the transformed latent variables $\mu(z_i)$. The parameter $\theta$ entails the parameters of the networks $\mu(\cdot)$ and $\sigma(\cdot)$. 

Deep latent Gaussian models are highly flexible density estimators. There exist many modified versions specific to other types of data. For example, for binary data, the Gaussian likelihood can be replaced by a Bernoulli likelihood. Next, we review how amortized inference is applied to this model class.

\paragraph{Variational Autoencoders}
Most commonly, VAEs refer to deep latent Gaussian models which are trained using inference networks.

VAEs employ two deep sets of neural networks: a top-down generative model as described above, mapping from the latent variables $\bmz$ to the data $\bmx$, and a bottom-up inference model which approximates the posterior  $p(\bmz|\bmx)$. Commonly, the corresponding neural networks are referred to as the \emph{generative network} and the \emph{recognition network}, or sometimes as decoder and encoder networks. 

In order to approximate the posterior, VAEs employ an amortized mean-field variational distribution:
$$q_\phi(\bmz|\bmx) = \prod_{i=1}^N q_\phi(z_i|x_i).$$ 
The conditioning on $x_i$ indicates that the local variational parameters associated with each data point are replaced by a function of the data. This amortized variational distribution is typically chosen as:

\begin{equation}
  q_\phi(z_i|x_i) = {\cal N}(z_i | \mu(x_i), \sigma^2(x_i)\mathbb{I}). \label{eq:vaepost}
\end{equation} 
Similar to the generative model, the variational distribution employs non-linear mappings $\mu(x_i)$ and $\sigma(x_i)$ of the data in order to predict the approximate posterior distribution of $x_i$. The parameter $\phi$ summarizes the corresponding neural network parameters.

The main contribution of \cite{kingma2013auto,rezende2014stochastic} was to derive a scalable and efficient training scheme for  deep latent variable models. During optimization, both the inference network and the generative network are trained jointly to optimize the ELBO.

The key to training these models is the reparameterization trick (Section \ref{sec:BBVI_rep}). We focus on the ELBO contribution form a single data point $x_i$. First, we draw $L$ samples $\epsilon_{(l,i)} \sim p(\epsilon)$ from a noise distribution.
We also employ a reparameterization function $g_{\phi}$, such that $z_{(i,l)} = g_{\phi}(\epsilon_{(l,i)},x_i)$ realize samples from the approximate posterior $q_{\phi}(z_i|x_i)$. For Eq.~\ref{eq:vaepost}, the most common reparametrization function takes the form $z_{(i,l)} = \mu(x_i) + \sigma(x_i) * \epsilon_{(i,l)}$, where $\mu(\cdot)$ and $\sigma(\cdot)$ are parameterized by $\phi$.
One obtains an unbiased Monte Carlo estimator of the VAE's ELBO by
\begin{align}
\hat{\mathcal{L}}(\theta, \phi, x_{i})  &=\label{eq:ELBOVAE}
 - D_{KL}(q_\phi(z_i |x_{i}) || p_\theta(z_i)) \\   &+ \frac{1}{L}\sum_{l=1}^L\log p_\theta(x_{i}|\mu(x_i) + \sigma(x_i) * \epsilon_{(i,l)}). \nonumber
\end{align}
This stochastic estimate of the ELBO can subsequently be differentiated with respect to $\theta$ and $\phi$ to obtain an estimate of the gradient.

The reparameterization trick also implies that the gradient variance is bounded by a constant 
\cite{rezende2014stochastic}. The drawback of this approach however is that we require the approximate posterior to be reparameterizable.  

\paragraph{A Probabilistic Encoder-Decoder Perspective}

The term \emph{variational autoencoder} arises from the fact that the joint training of the generative and recognition network resembles the structure of autoencoders, a class of unsupervised, deterministic models. 
Autoencoders are deep neural networks that are trained to reconstruct their inputs as closely as possible. Importantly, the neural networks involved in autoencoders have an hourglass structure, meaning that there is a small number of units in the inner layers that prevent the neural network from learning the trivial identity mapping. This 'bottleneck' forces the network to learn a useful and compact representation of the data.

In contrast, VAEs are probabilistic models, but they have a close correspondence to classical autoencoders. It turns out that the hidden variables of the VAE can be thought of as the intermediate representations of the data in the bottleneck of an autoencoder. During VAE training, one injects noise into this intermediate layer, which has a regularizing effect. In addition, the KL divergence term between the prior and the approximate posterior forces the latent representation of the VAE to be close to the prior, leading to a more homogeneous distribution in latent space that generalizes better to unseen data. When the variance of the noise is reduced to zero and the prior term is omitted, the VAE becomes a classical autoencoder.

\subsection{Advancements in VAEs}
\label{sec:advanceVAE}
Since the proposal of VAEs, an ever-growing number of extensions have been proposed. 
While exhaustive coverage of the topic would require a review article in its own right, we summarize a few important extensions. While several model extensions of the VAE have been proposed, this review puts a bigger emphasis on inference procedures. We will report on extensions that modify the variational approximation $q_\phi$, the model $p_\theta$, and finally discuss the dying units problem when the posterior of some latent units remains close to the prior during the optimization.
\vspace{-5pt}
\paragraph{Flexible Variational Distributions $\mathbf{q_\phi}$}
Traditional VI, including VAE training, relies on parametric inference models.
The approximate posterior $q_\phi$ can  be an explicit parametric distribution, such as a Gaussian or discrete distribution \cite{rolfe2016discrete}. We can use more flexible distributions, for example by transforming a simple parametric distribution. Here, we review VAE with implicit distributions, normalizing flow, and importance weighted VAE. Using more flexible variational distributions reduces not only the approximation error but also the amortization error, i.e. the error introduced by replacing the local variational parameters by a parametric function \cite{cremer2018inference}.

Implicit distributions can be used in VI since a closed-form density function is not a strict requirement for the inference model; all we need is to be able to sample from it.
As detailed below, their reparameterization gradients can still be computed. In addition to the standard reparameterization approach, the entropy contribution to the gradient has to be estimated. 
Implicit distributions for VI is an active area of research \cite{li2016wild,huszar2017variational, karaletsos2016adversarial,mohamed2016learning,mescheder2017adversarial, li2017approximate,liu2016two,titsias2017learning,wang2016learning}. 

VI requires the computation of the log density ratio $\log p(\bmz) - \log q_{\phi}(\bmz|\bmx)$.  When $q$ is implicit, the standard training procedure faces the problem that log density ratio is intractable. In this case, one can employ a Generative Adversarial Networks (GAN)  \cite{goodfellow2014generative} style discriminator $T$ that discriminates the prior from the variational distribution, $T(\bmx, \bmz) = \log q_{\phi}(\bmz|\bmx) - \log p(\bmz)$ \cite{mescheder2017adversarial,li2016wild}. This formulation is very general and can be combined with other ideas, for example a  hierarchical structure \cite{tran2017hierarchical,yin2018semi} .

Normalizing flow \cite{ding2014nice,rezende2015variational, dinh2017density, chen2016variational, kingma2016Improving} presents another way to utilize flexible variational distributions.
 The main idea behind normalizing flow is to transform a simple (e.g. mean field) approximate posterior $q(\bmz)$ into a more expressive distribution by a series of successive invertible transformations. 
 
To this end, we first draw a random variable $z \sim q(\bmz)$, and transform it using an invertible, smooth function $f$. Let $z' = f(z)$. Then, the new distribution is
\begin{equation}
q(z') = q(z)|\frac{\partial f^{-1}}{\partial z'}| = q(z)|\frac{\partial f}{\partial z'}|^{-1}. 
\end{equation}
It is necessary that we can compute the determinant, since the variational approach requires us to also estimate the entropy of the transformed distribution.
By choosing the transformation function $f$ such that $|\frac{\partial f}{\partial z'}|$ is easily computable, this normalizing flow constitutes an efficient method to generate multimodal distributions from a simple distribution. As variants, linear time-transformations, Langevin and Hamiltonian flow  \cite{rezende2015variational}, as well as inverse autoregressive flow  \cite{kingma2016Improving}  and autoregressive flow  \cite{chen2016variational} have been proposed. 

Normalizing flow and the previously mentioned implicit distribution share the common idea of using transformations to transform simple distributions into more complicated ones. A key difference is that for normalizing flows, the density of $q(z)$ can be estimated due to an invertible transformation function.

One final approach that utilizes flexible variational distributions is the importance weighted variational autoencoder (IWAE) which was originally proposed to tighten the variational bound \cite{burda2015importance} and can be reinterpreted to sample from a more flexible distribution \cite{cremer2017reinter}.
IWAEs require $L$ samples from the approximate posterior which are weighted by the ratio 
\begin{equation}
\hat{w}_l = \frac{w_l}{\sum_{l=1}^L w_l}, \text{where }  w_l = \frac{p_\theta(x_{i}, z_{(i,l)})}{q_\phi(z_{(i,l)}|x_{i})}.
\end{equation}
The authors show that the more samples $L$ are evaluated, the tighter the variational bound becomes, implying that the true log likelihood is approached in the limit $L \rightarrow \infty$. A reinterpretation of IWAEs, suggests that they are identical to VAEs but sample from a more expressive distribution which converges pointwise to the true posterior as $L \rightarrow \infty$ \cite{cremer2017reinter}.  As IWAEs introduce a biased estimator, additional steps to obtain potentially better variance-bias trade-offs can be taken, such as in\cite{nowozin2018debiasing,rainforth2018onnesting,rainforth2018tighter} . 

\paragraph{Modeling Choices of $\mathbf{p_\theta}$} 

Modeling choices affect the performance of deep latent Gaussian models.
In particular improving the prior model in VAEs can lead to more interpretable
fits and better model performance.
\cite{johnson2016structured} proposed a method to utilize a structured prior for VAEs, combining the advantages of traditional graphical models and inference networks. 
These hybrid models overcome the intractability of non-conjugate priors and likelihoods  by learning variational parameters of conjugate distributions with a recognition model. This allows one to approximate the posterior conditioned on the observations while maintaining conjugacy. As the encoder outputs an estimate of natural parameters, message passing, which relies on conjugacy, is applied to carry out the remaining inference.

Other approaches tackle the drawback of the standard VAE which is the assumption that the likelihood factorizes over dimensions. This can be a poor approximation, e.g., for images,
for which a structured output model works better.
The Deep Recurrent Attentive Writer \cite{gregor2015draw} relies on a recurrent structure that gradually constructs the observations while automatically focusing on regions of interest.
In comparison, PixelVAE \cite{gulrajani2017pixelvae} tackles this problem by modeling dependencies between pixels within an image, using a conditional model that factorizes as $p_\theta(x_i|z_i) = \prod_j p_\theta(x_i^j|x_i^1,...x_i^{j-1},z_i)$, where $x_i^j$ denotes the $j$th dimension of observation $i$. The dimensions are generated in a sequential fashion, which accounts for local dependencies within the pixels of an image. The expressiveness of the modeling choice comes at a cost. If the decoder is too strong, the inference procedure can fail to learn an informative posterior  \cite{ chen2016variational}. This problem, known as the dying units problem, will be discussed in the paragraph below.


\paragraph{The Dying Units Problem}

Certain modeling choices and parameter configurations impose problems in VAE training, such that learning a good low-dimensional representation of the data fails. A prominent such problem is known as
the \emph{dying units problem}.
As detailed below, two main effects are responsible for this phenomenon: a too powerful decoder, and the KL divergence term.

In some cases, the expressiveness of the decoder can be so strong, that some dimensions of the $\bmz$ variables are ignored, i.e. it might model  $p_\theta(\mathbf{x}|\mathbf{z})$  independently of $\mathbf{z}$. In this case the true posterior is the prior \cite{chen2016variational}, and thus the variational posterior tries to match the prior in order to satisfy the KL divergence in Eq. \ref{eq:ELBOVAE}.  
Lossy variational autoencoders \cite{chen2016variational} circumvent this problem by conditioning the decoding distribution for each output dimension on partial input information. For example, in the case of images, the likelihood of a given pixel is only conditioned on the values of the immediate surrounding pixels and the global latent state. 
This forces the distribution to encode global information in the latent variables.

The KL divergence contribution to the VAE loss may exacerbate this problem. To see why, we can rewrite the ELBO as a sum of two KL divergences $\hat{\mathcal{L}}(\theta, \phi, x_{i})=- D_{KL}(q_\phi(z |x_{i}) || p_\theta(z))-D_{KL}(p(x_i) || p_\theta(x_{i}| z)) +C$. If the model is expressive enough, the model is able to render the second term zero (independent of the value of $\bmz$). In this case, in order to also satisfy the first term, the inference model places its probability mass to match the prior \cite{zhaoSE17a}, failing to learn a useful representation of the data.  
Even if the decoder is not strong, the problem of dying units may arise in the early stages of the optimization where the approximate posterior does not yet carry relevant information about the data \cite{bowman2016generating}. This problem is more severe when the dimension of $z$ is high. In this situation, units are regularized towards the prior and might not be reactivated in the later stages of the optimization  \cite{sonderby2016train}. To counteract the early influence of the KL constraint, an annealing scheme can be applied to the KL divergence term during training \cite{sonderby2016train}.

\vspace{-5pt}
\section{Discussion}
\label{sec:link}
We have summarized recent advancements in variational inference. Here we outline some selected active research directions and open questions, including, but not limited to: theory of VI, VI and policy gradients, VI for deep learning (DL), and automatic VI. 

\paragraph{Theory of VI}
Despite progress in modeling and inference, few authors address theoretical aspects of VI \cite{nakajima2007variational,lacoste2011approximate,wang2018frequentist}. 
One important direction is quantifying the approximation errors involved when replacing a true posterior with a simplified variational distribution \cite{nakajima2007variational}. A related problem is the predictive error, e.g., when approximating the marginalization involved in a Bayesian predictive distribution using VI.

We also conjecture that VI theory could profit from a deeper connection with information theory. This was already exemplified in \cite{tanaka2000information,tanaka1999theory}.
Information theory also inspires the development of new models and inference schemes
\cite{barber2003algorithm,tishby2000information,alemi2017deep}. For example, the information bottleneck \cite{tishby2000information} has recently led to the deep variational information bottleneck \cite{alemi2017deep}. We expect more interesting results to come from fusing these two lines of research.

\paragraph{VI and Deep Learning} 
 
Despite its recent successes in various areas, deep learning still suffers from a lack of principled uncertainty estimation, a lack in interpretability of its feature representations, and difficulties in including prior knowledge. Bayesian approaches, such as Bayesian neural networks \cite{neal2012bayesian} and  variational autoencoders (as reviewed in Section \ref{sec:DAIforDL}),  are improving all these aspects. Recent work aims at using interpretable probabilistic models as priors for VAEs \cite{johnson2016structured,saeedi2017multimodal,krishnan2015deep,deng2017factorized}. In such models, VI is an essential component. Making VI computationally efficient and easy to implement in Bayesian deep architectures is becoming an important research direction \cite{gal2016uncertainty}.  

\paragraph{VI and Policy Gradients}
Policy gradient estimation is important for reinforcement learning (RL) \cite{sutton1998reinforcement} and stochastic control. The technical challenges in these applications are similar to VI \cite{levine2018reinforcement, levine2013variational,schulman2015trust,liu2017stein,wang17Variational} (See Appendix \ref{sec:App_PolicyGradient}). As an example, SVGD has been applied in the RL setting as the Stein policy gradient \cite{liu2017stein}. The application of VI in RL is currently an active area of research.

\paragraph{Automatic VI} 
 Probabilistic programming allows practitioners to quickly implement and revise models without having to worry about inference. The user is only required to specify the model, and the inference engine will automatically perform the inference. Popular probabilistic programming tools include but are not limited to: \textit{Stan} \cite{carpenter2016stan}, which covers a large range of advanced VI and MCMC methods, \textit{Infer.Net} \cite{InferNET14}, which is based on variational message passing and EP,  \textit{Automatic Statistician} \cite{ghahramani2015probabilistic}  and \textit{Anglican} \cite{tolpin2016design}, which mainly rely on sampling methods,  \textit{Edward} \cite{tran2016edward}, which supports BBVI as well as Monte Carlo sampling, and \textit{Zhusuan} \cite{shi2017zhusuan}, which features VI for Bayesian Deep learning.  The longstanding goal of these tools is to change the research methodology in probabilistic modeling, allowing users to quickly revise and improve models 
and to make them accessible to a broader audience.

Despite current efforts  to make  VI more accessible to practitioners, its usage is still not straightforward for non-experts. 
For example, manually identifying posterior symmetries and  breaking these symmetries is necessary to work with Infer.Net.
Furthermore,  variance reduction methods such as control variates can drastically accelerate convergence, but a model specific design of control variates is needed to obtain the best performance. At the time of writing, these problems are not yet addressed in current probabilistic programming toolboxes. We believe these and other directions are important to advance the impact of probabilistic modeling in science and technology.

\section{Conclusions}
\label{sec:conclusions}
In this paper, we review the recent major advances in variational inference  from four perspectives: scalability, generality, accuracy,  and amortized inference. The advancement of variational inference theory and the adoption of approximate inference in new machine learning models are developing rapidly. 
Although this field has grown in recent years, it remains an open question how to make VI more efficient, more accurate, and easier to use for non-experts. Further development, as discussed in the previous section, is needed.

\ifCLASSOPTIONcompsoc
  \section*{Acknowledgments}
\else
  \section*{Acknowledgment}
\fi

The authors would like to thank Sebastian Nowozin, Francisco Ruiz, Tianfan Fu, Robert Bamler, and especially Andrew Hartnett and  Yingzhen Li for comments and discussions that greatly improved the manuscript. 

\ifCLASSOPTIONcaptionsoff
  \newpage
\fi

\bibliography{ref}
\bibliographystyle{plain}

\vspace{-23pt}
\begin{IEEEbiography}[{\includegraphics[width=1in,height=1.25in,clip,keepaspectratio]{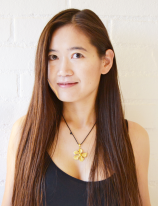}}]{Cheng Zhang} is a researcher at Microsoft Research, Cambridge, UK.  Previously, she was a postdoctoral research associate at Disney Research, Pittsburgh, USA. She obtained her master's degree and her Ph.D. degree from KTH, Stockholm, Sweden.  Her research interest lies in probabilistic modeling, representation learning, causality, and approximate inference, as well as their applications in e.g. Computer Vision and Healthcare.
\end{IEEEbiography}
\vspace{-25pt}
\begin{IEEEbiography}[{\includegraphics[width=1in,height=1.25in,clip,keepaspectratio]{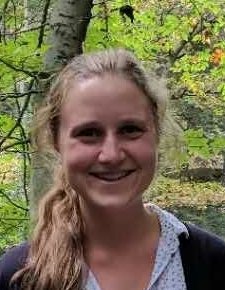}}]{Judith B\"utepage} holds a Bsc degree in Cognitive Science from the University of Osnabru\"ck. She
is a doctoral student at the Department of Robotics, Perception, and Learning (RPL)
  at KTH in Stockholm, Sweden.  Her research interests lie in the development of Bayesian latent variable models and their application to problems in Robotics and Computer Vision.
\end{IEEEbiography}
\vspace{-25pt}
\begin{IEEEbiography}[{\includegraphics[width=1in,height=1.25in,clip,keepaspectratio]{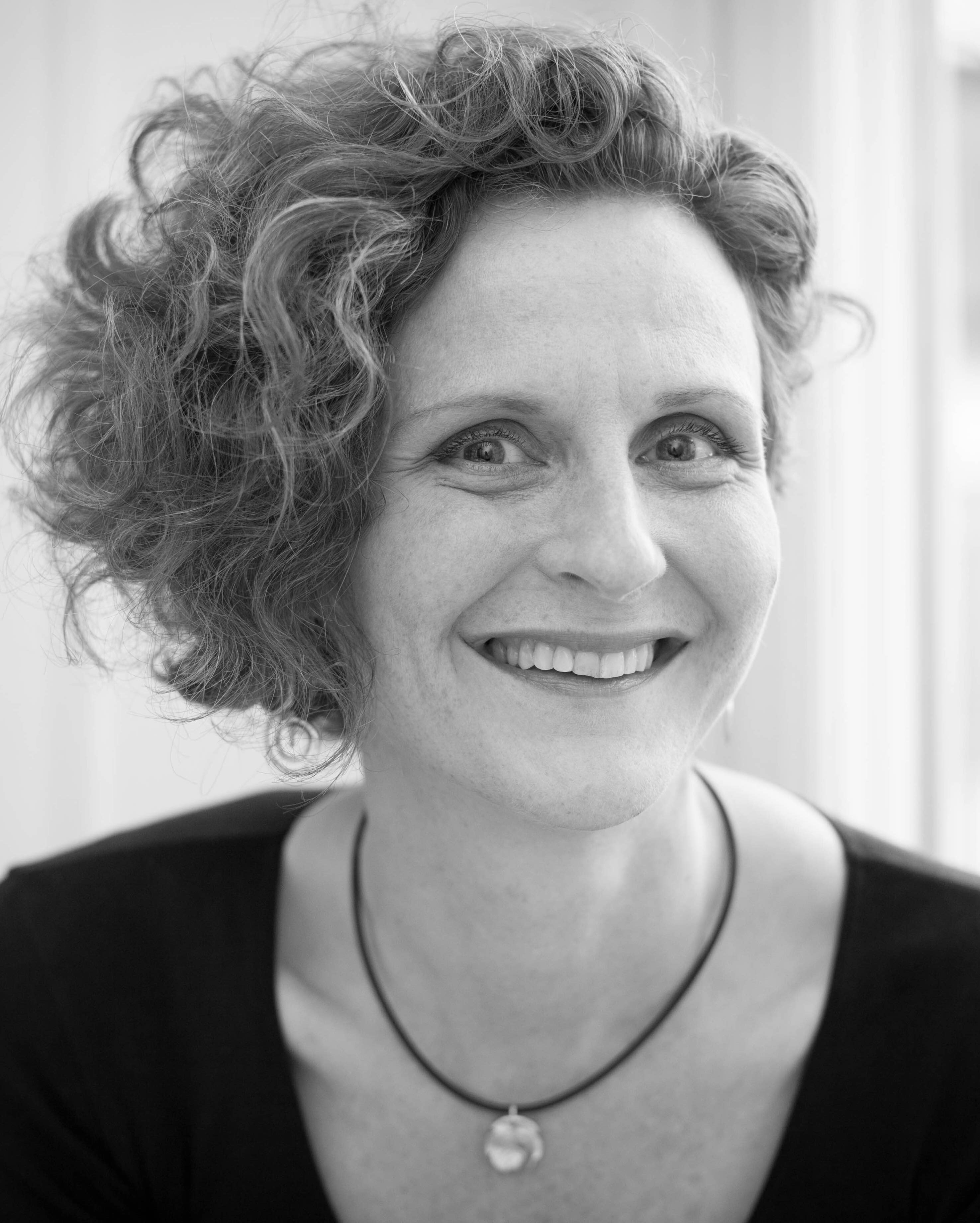}}]{Hedvig Kjellstr\"om}
 is a Professor 
  and the head of the Department of Robotics, Perception, and Learning (RPL)
  at KTH in Stockholm, Sweden. Her present
  research focuses on the modeling of perception and production of human
  non-verbal communicative behavior and activity. 
In 2010, she was awarded the Koenderink Prize for fundamental
  contributions in Computer Vision.  She has written around 85 papers,
  and is an Associate Editor for IEEE TPAMI and IEEE RA-L.   
\end{IEEEbiography}
\vspace{-25pt}
\begin{IEEEbiography}[{\includegraphics[width=1in,height=1.25in,clip,keepaspectratio]{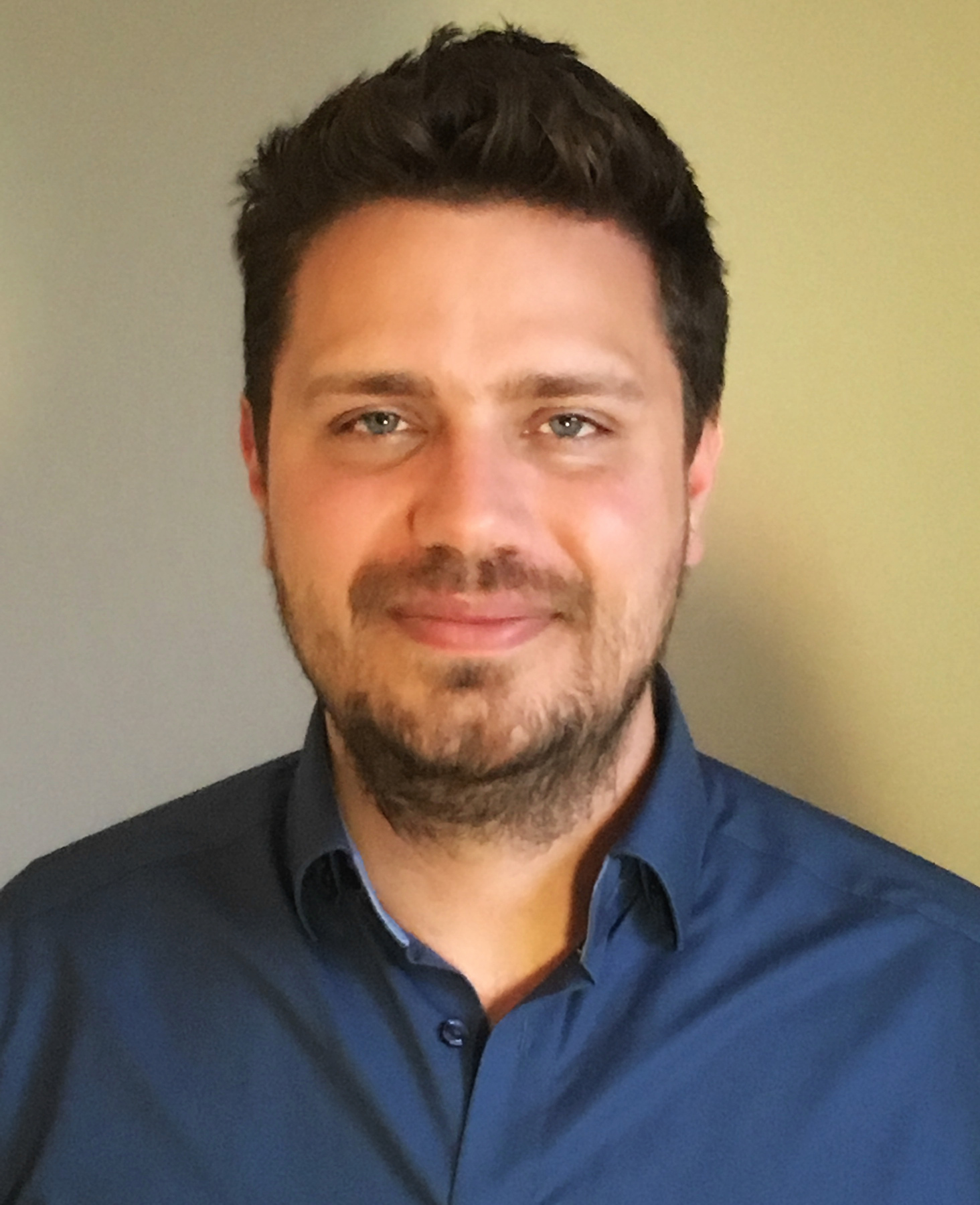}}]{Stephan Mandt}
is an Assistant Professor at the University of California, Irvine. Previously, he led the statistical machine learning labs at Disney Research Pittsburgh and Los Angeles as Research Scientist and Senior Research Scientist, respectively. Stephan was a postdoctoral researcher with David Blei at Columbia and Princeton University. He holds a Ph.D. in theoretical physics from the University of Cologne, supported by the German National Merit Foundation. His interests include scalable probabilistic modeling, variational inference, Bayesian deep learning, and applications in the sciences and digital media.
\end{IEEEbiography}

 \clearpage
 \appendices
 \section{ }
\subsection{ELBO and KL}
\label{sec:ELBO_KL}

We show that the difference between the marginal likelihood $\log p(\bmx)$ and the ELBO $\mathcal{L}$ is the KL divergence between the variational distribution $q(\bmz; \bm{\lambda})$ and the target distribution $p(\bmx,\bmz)$: 
\begin{equation}
\begin{aligned}
&\log p(\bmx)-\mathcal{L} =\log p(\bmx)-\E_{q(\bmz ; \bm{\lambda})} \left[ \log \frac{p(\bmx,\bmz)}{q(\bmz; \bm{\lambda})}  \right] \\
&= \E_{q(\bmz; \bm{\lambda})}  \left[\log p(\bmx) - \log \frac{p(\bmx,\bmz)}{q(\bmz; \bm{\lambda})}  \right] \\
&= -\E_{q(\bmz)}\left[ \log \frac{p(\bmz|\bmx)}{q(\bmz)}  \right] 
=D_{\text{KL}} (q || p).  
\label{eq:KLdiveElbo}
\end{aligned}
\end{equation}
With this equivalence, the ELBO $\mathcal{L}$ can be derived using either Jensen's inequality as in Eq. \ref{eq:ELBO}, or using the KL divergence as $\mathcal{L} = \log p(x) - D_{\text{KL}} (q || p)$.


\subsection{Conjugate Exponential family}
\label{sec:ExpFam}

Many probabilistic models involve exponential family distributions. 
A random variable $x$ is distributed according to a member of the exponential family if its probability distribution  takes the form
$$ p(x|\theta) = h(x)exp(\eta(\theta)t(x) - a(\eta(\theta))),$$
where $\theta$ is a vector of parameters, $\eta(\cdot)$ is the natural parameter, and $t(\cdot)$ are the sufficient statistics. Furthermore, $h(\cdot)$ is the base measure and $a(\cdot)$ is the log-normalizer. Many distributions fall into this class.

In the context of Bayesian statistics, certain exponential family distributions are conjugate pairs. A likelihood and prior distribution are a conjugate pair if
the corresponding posterior distribution is in the same family as the prior. Examples for conjugate pairs include a Gaussian distribution with a Gaussian prior, a Poisson distribution with a gamma prior, or a multinomial distribution with a Dirichlet prior.

\subsection{Variational Message Passing}
\label{sec:VMP}

Winn \textit{et. al.}  formulate MFVI in a message passing manner \cite{winn2005variational}. 
MFVI provides a method to update the latent variables of the variational distribution sequentially, as shown in Equation \ref{eq:update}. 
In a Bayesian network, the update for each node only requires information from the nodes in its Markov blanket, which includes this node's parents, children, and co-parents of its children,
\begin{equation}
\begin{aligned}
q^*( z_j) & \propto \exp( \E_{q(\bmz_{\neg j})} \ [ \log p(z_j | \bm{pa}_j)]  \\
&+ \sum_{c_k \in \bm{ch}_j}  \E_{q(\bmz_{\neg j})} \ [ \log p(c_k | \bm{pa}_k)]   ),
\end{aligned}
\label{eq:update_MB}
\end{equation}
where $\bm{pa}_j$ indicates the set of parent nodes of $z_j$, $\bm{ch}_j$ includes the set of the child nodes of $z_j$, and $c_k$ indicates the $k$th child node. $\bm{pa}_k$ indicates the set of parent nodes of $c_k$. Hence, the update of one latent variable only depends on its parents, children, and its children's co-parents.

If we further assume that the model is conjugate-exponential, see Section \ref{sec:ExpFam}, a latent variable can be updated by receiving all messages from its parents and children. Here, each child node has already received messages from its co-parents. Thus, to update each node, only nodes in this node's Markov blanket are involved. Finally, $z_j$ is updated with the following three steps: a) receive messages from all parents and children $m_{\bm{pa}_j\rightarrow z_j} = \langle t_{\bm{pa}_j} \rangle $,  $m_{c_k\rightarrow z_j} = \tilde{\eta}_{c_kz_j} ( \langle t_{c_k} \rangle, \{m_{i \rightarrow c_k}\}_{i \in \bm{pa}_k}) $; b) update $z_j$'s natural parameter $\eta_{z_j}$; c) update the expectation of $z_j$'s sufficient statistic $\langle t(z_j) \rangle$. 

Variational message passing provides a general message passing formulation for the MFVI. It enjoys all the properties of MFVI, but can be used in large scale Bayesian networks and can be automated easily. Together with EP, it forms the basis for the popular probabilistic programming tool Infer.Net \cite{InferNET14}. 

\subsection{Natural Gradients and SVI}
\label{sec:svi_naturalgradient}
We use the model example as shown in Figure \ref{fig:generativemodel} and assume that the true posterior of the global variable is the in exponential family:
\begin{align*}
&p(\theta |  x,\lz, \alpha) \\
&= h(\theta) \exp \Big( 
\eta_g(x,\lz, \alpha)^T t(\theta) - a_g\big(\eta_g(x,\lz, \alpha)^T t(\theta)\big)
\Big).
\end{align*}
We also assume that the variational distribution is in the same family:
\begin{align*}
&q(\theta |  \gamma) = h(\theta) \exp \Big( 
\gamma^T t(\theta) - a_g(\gamma)
\Big).
\end{align*}
Recall that $\gamma$ is the variational parameter estimating the global variable $\theta$. The subscript $g$ in $\eta_g$ and $a_g$ denotes that these are the natural parameter and log-normalizer of the global variable.
The natural gradient of a function $f(\gamma)$ is given by $\hat{\nabla}_\gamma f(\gamma) = G(\gamma)^{-1}\nabla_\gamma f(\gamma)$, where $G(\gamma)$ is the Fisher information matrix.

\cite{hoffman13} showed that the ELBO has a closed-form solution in terms of its variational parameters $\gamma$:
\begin{align}
\hat{\mathcal{L}}(\gamma) = & \label{Eq:ELBOSVING} \\ \E_q \nonumber & \left[\eta_g(\bmx,\bmz,\alpha) \right]\nabla_\gamma a_g(\gamma) - \gamma^T\nabla_\gamma a_g(\gamma) + a_g(\gamma) + c. 
\end{align}
The constant $c$ contains all those terms that are independent of $\gamma$. The gradient of Equation \ref{Eq:ELBOSVING} is given by  
\begin{align}
 \nabla_\gamma \hat{\mathcal{L}}(\gamma) = \nabla^2_\gamma a_g(\gamma) ( \E_q \nonumber & \left[\eta_g(\bmx,\bmz,\alpha) \right] - \gamma). 
\end{align}
Importantly, when $q(\gtht|\gamma)$ is in the exponential family, then it holds that $G(\gamma) =  \nabla^2_\gamma a_g(\gamma)$. Thus, the natural gradient simplifies to
\begin{align}
 \hat{\nabla}_\gamma  \hat{\mathcal{L}}(\gamma) =  \E_q \nonumber & \left[\eta_g(\bmx,\bmz,\alpha) \right] - \gamma. 
\end{align}
Hence, the natural gradient has a simpler form than the regular gradient.

Following the natural gradient has the advantage that we do not optimize in the Euclidean space, which is often not able to represent distances between distributions, but in Riemann space, where distance is defined by the KL divergence, i.e. distance between distributions. More information about the advantages of using natural gradients can be found in \cite{amari1998natural}.

\subsection{Rao-Blackwell Theorem}
\label{sec:raoBlackwell}

Rao-Blackwellization is used in multiple VI methods for variance reduction such as in BBVI\cite{ranganath2014black}. In general, the Rao-Blackwell Therorem \cite{hogg1995introduction} states the following: 
Let $\hat \theta$ be an estimator of parameter $\theta$ with $\E(\hat \theta ^2) < \infty$ for all $\theta$. Suppose that $t$ is a sufficient statistic for $\theta$, and let $\theta^* = \E(\hat \theta |t )$. Then for all $\theta$,
$$
E(\theta^* -\theta) ^2 \le E(\hat \theta - \theta) ^2.
$$
The inequality is strict unless $\hat \theta$ is a function of $t$.
This implies that the conditional estimator $\theta^* = \E(\hat \theta |t)$,  conditioned on the sufficient statistics, is a better estimator than any other estimator $\hat \theta$.

\subsection{Physics Notations}
\label{sec:Physics_Notations}
In order to facilitate the comprehension of the older literature on VI, we introduce some notation commonly used by the physics community \cite{opper2001advanced}.
Distributions are commonly denoted by capital letters $P$ and $Q$.
We can write the KL divergence as:
\begin{equation*}
KL(Q||P) = \log Z + \E_Q[\log \, P] - \mathbb{H}[Q],
\end{equation*}
which corresponds to Equation \ref{eq:KLdiveElbo}.  Here, $\mathbb{H}$ denotes the entropy of a distribution.
In the physics community, $-\log Z$ is called free energy. $Z$ is the commonly the marginal likelihood in machine learning, and often called the partition function in physics. $E_Q[\log \,P]$ is called the variational energy and $F[Q] = E[\log \,P] - \mathbb{H}[Q]$ is the variational free energy which correspond to the negative ELBO, $F[Q] = - \mathcal{L}$.

\subsection{Policy Gradient Estimation as VI}
\label{sec:App_PolicyGradient}

Reinforcement learning (RL) with policy gradients can be formulated as a VI problem \cite{liu2017stein}. 
In RL, the objective is to maximize the expected return
\begin{equation}
J(\theta) = J\big(\pi(a| s; \theta) \big) 
= \E_ {\bm{s},\bm{a}} \left[\sum_{t=0} ^\infty \gamma ^t r(s_t, a_t) \right],
\end{equation}
where $\pi(a| s; \theta)$ indicates the policy parameterized by $\theta$, $r$ is a scalar reward for being in state $s_t$ and performing action $a_t$  at time $t$, and  $\gamma$ is the discount factor.
The policy optimization can be formulated as a VI problem by using $q(\theta)$ -- a variational distribution on $\theta$ -- to 
maximize $\E_{q(\theta)} [J(\theta)]$. Using a max-entropy regularization, the optimization objective is
\begin{equation}
\mathcal{L} = \E_{q(\theta)}[ J(\theta)] + \alpha H (q(\theta)).
\end{equation}
This objective is the identical to the ELBO for VI.

\end{document}